\documentclass[letterpaper]{article} 
\usepackage{aaai2026}  
\usepackage{times}  
\usepackage{helvet}  
\usepackage{courier}  
\usepackage[hyphens]{url}  
\usepackage{graphicx} 
\usepackage{natbib}  
\usepackage{caption} 
\usepackage{algorithm}
\usepackage{algorithmic}

\usepackage{newfloat}
\usepackage{listings}
\DeclareCaptionStyle{ruled}{labelfont=normalfont,labelsep=colon,strut=off} 
\floatstyle{ruled}
\newfloat{listing}{tb}{lst}{}
\floatname{listing}{Listing}
\usepackage[utf8]{inputenc} 
\usepackage[T1]{fontenc}    
\usepackage{url}            
\usepackage{booktabs}       
\usepackage{amsfonts}       
\usepackage{nicefrac}       
\usepackage{microtype}      
\usepackage{enumitem}
\usepackage{algorithm}
\usepackage{algorithmic}
\usepackage{amsmath}
\usepackage{xcolor}
\usepackage[capitalize,noabbrev]{cleveref}
\usepackage[page,header]{appendix}
\usepackage{titletoc}
\usepackage{minitoc}
\usepackage{subfigure}
\usepackage{lipsum}
\usepackage{ntheorem}

\newtheorem{theorem}{Theorem}

\newtheorem{lemma}{Lemma}
\newtheorem{corollary}{Corollary}
\newtheorem{assumption}{Assumption}

\newtheorem{definition}{Definition}
\newtheorem*{remark}{Remark}

\newtheorem*{proof}{Proof}

\newcommand{\bs}[1]{\boldsymbol{#1}}

\title{Understanding Generalization of Federated Learning: \\ the Trade-off between Model Stability and Optimization}
\author{
    Dun Zeng\equalcontrib\textsuperscript{\rm ,1},
    Zheshun Wu\equalcontrib\textsuperscript{\rm ,3}, 
    Shiyu Liu\textsuperscript{\rm 1}, 
    Yu Pan\textsuperscript{\rm 3}, 
    Xiaoying Tang\textsuperscript{\rm 4}, 
    Zenglin Xu\textsuperscript{\rm 2}
}
\affiliations{
    \textsuperscript{\rm 1}University of Electronic Science and Technology of China\\

    \textsuperscript{\rm 2}Fudan University, \textsuperscript{\rm 3}Harbin Institute of Technology, Shenzhen\\
    \textsuperscript{\rm 4}Chinese University of Hong Kong, Shenzhen \\
    zengdun@std.uestc.edu.cn, wuzhsh23@gmail.com, shiyu.liu@foxmai.com\\
    iperryuu@gmail.com, tangxiaoying@cuhk.edu.cn, zenglinxu@fudan.edu.cn
}

\usepackage{bibentry}

\begin{document}

\maketitle

\begin{abstract}
Federated Learning (FL) is a distributed learning approach that trains machine learning models across multiple devices while keeping their local data private. However, FL often faces challenges due to data heterogeneity, leading to inconsistent local optima among clients. These inconsistencies can cause unfavorable convergence behavior and generalization performance degradation. 
Existing studies often describe this issue through \textit{convergence analysis} on gradient norms, focusing on how well a model fits training data, or through \textit{algorithmic stability}, which examines the generalization gap. However, neither approach precisely captures the generalization performance of FL algorithms, especially for non-convex neural network training.
In response, this paper introduces an innovative generalization dynamics analysis framework, namely \textit{Libra}, for algorithm-dependent excess risk minimization, highlighting the trade-offs between model stability and gradient norms. We present Libra towards a standard federated optimization framework and its variants using server momentum. Through this framework, we show that larger local steps or momentum accelerate convergence of gradient norms, while worsening model stability, yielding better excess risk. Experimental results on standard FL settings prove the insights of our theories. These insights can guide hyperparameter tuning and future algorithm design to achieve stronger generalization.
\end{abstract}


\section{Introduction}

Federated Learning (FL) has become a new solution for training AI models on distributed and private data~\citep{mcmahan2017communication, kairouz2021advances}. Traditional centralized learning approaches require data to be gathered in one place, posing risks to privacy and security~\citep{voigt2017eu, li2019impact, zhang2023survey}. In response, FL trains models directly on local devices, keeping raw data decentralized. This paradigm has enabled many applications in fields like healthcare~\citep{antunes2022federated}, finance~\citep{long2020federated}, and IoT~\citep{nguyen2021federated}, where protecting sensitive data is crucial. However, FL often faces significant performance challenges due to heterogeneous data in real-world scenarios. Many experimental and theoretical results of FL have demonstrated that the heterogeneity issues degrade the FL performance on both training behavior and testing accuracy~\citep{huang2023federated}. Therefore, numerous studies in FL have been proposed to alleviate the negative impacts of heterogeneity.

Most existing studies of FL emphasize convergence to empirically optimal solutions on the training data~\citep{reddi2020adaptive, wang2022communication, li2019convergence}, while often overlooking the generalization properties of the learned models.
In response, recent works have tried to address the generalization aspects of FL~\citep{sun2024understanding, init2024understanding}, bringing greater attention to this critical issue within the FL community.
These generalization studies are largely grounded in the theory of \textit{Uniform Stability}~\citep{hardt2016train}, a well-established framework in classical learning theory. However, uniform stability often fails to provide meaningful insights in general non-convex settings\citep{chen2018stability, charles2018stability, zhou2018generalization, li2019generalization}, which are typical of modern neural networks. Therefore, understanding generalization in FL under non-convex regimes is essential for its real-world applications.
One promising direction for this is the analysis of \textit{generalization dynamics}~\citep{teng2021towards}, which evaluates how the generalization gap evolves during training. While uniform stability primarily quantifies the stability of empirical loss, it often overlooks the role of non-zero gradient norm dynamics, making it less suitable for capturing the behavior of neural networks during training. Conversely, convergence analysis in non-convex FL primarily focuses on the convergence of gradient norms, while neglecting generalization considerations altogether.
This disconnect reveals a fundamental limitation: neither stability nor convergence analysis alone can adequately explain the generalization behavior of non-convex models. This motivates the need for a unified analytical framework that jointly captures convergence and stability, enabling a deeper understanding of generalization dynamics in FL model training.

\begin{figure*}[t]
\centering
\subfigure[Under-fitting]{
\begin{minipage}[t]{0.3\textwidth}
\centering
\includegraphics[width=\linewidth]{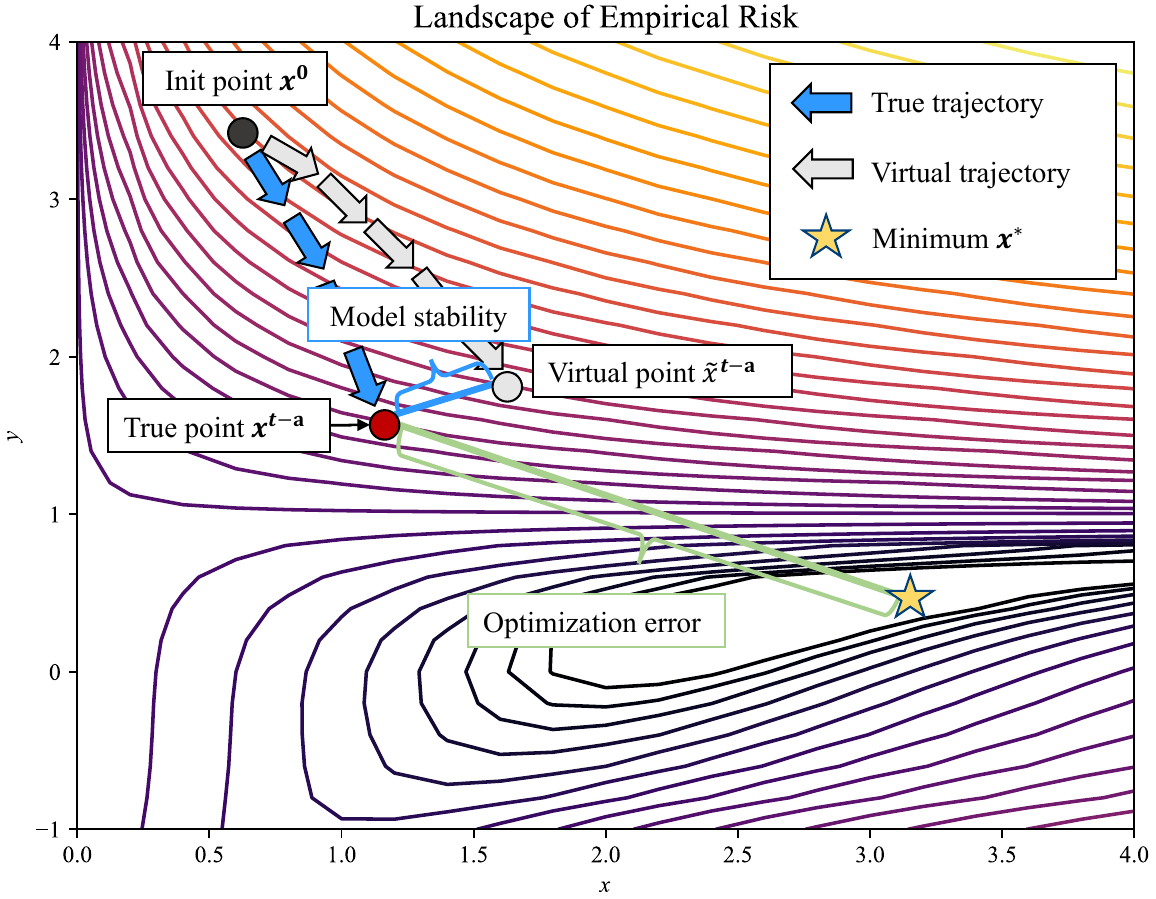}
\end{minipage}}
\subfigure[Benign-fitting]{
\begin{minipage}[t]{0.3\textwidth}
\includegraphics[width=\linewidth]{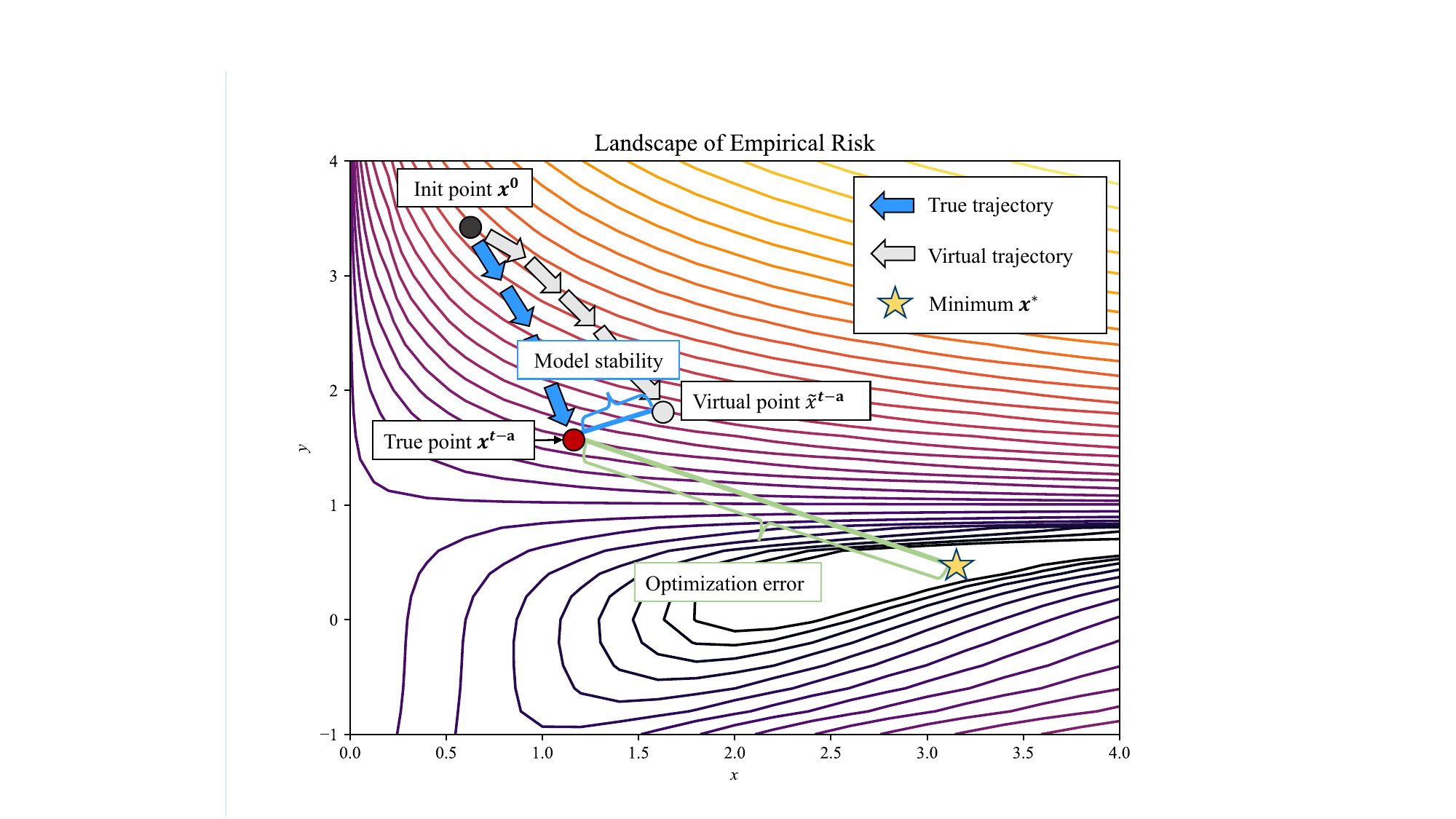}
\end{minipage}}
\subfigure[Over-fitting]{
\begin{minipage}[t]{0.3\textwidth}
\centering
\includegraphics[width=\linewidth]{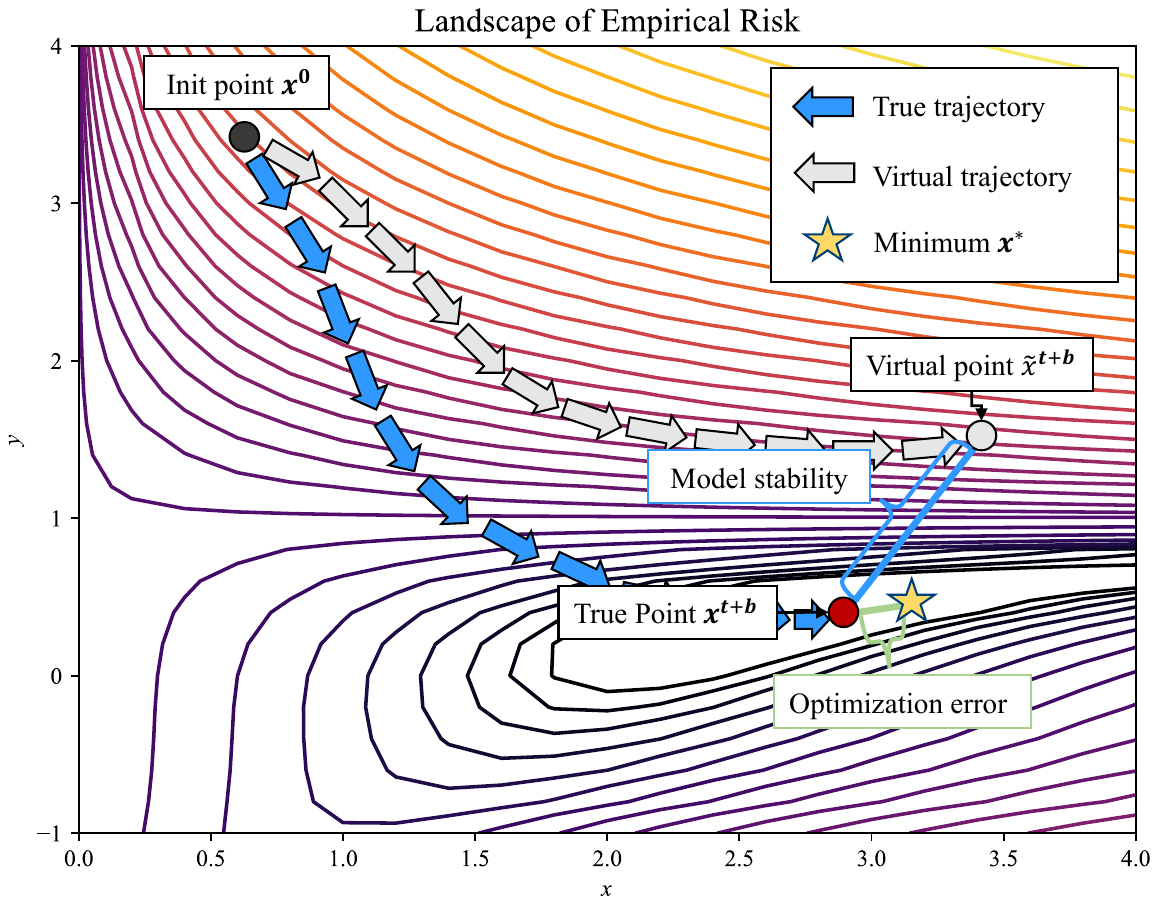}
\end{minipage}}
\caption{\textbf{Libra: understanding generalization dynamics via jointly analyzing model stability and optimization error.} In Corollary~\ref{corollary:excess_dynamics}, we show that the excess risk is roughly bounded by the sum of optimization error (green line) and model stability (blue line). In a training process, the green line shrinks, and the blue line expands at different speeds. At an early stage, the optimization error dominates the excess risk bound, indicating under-fitting. When the model stability dominates the excess risk, the model is over-fitting.
Libra can identify the algorithm-dependent factors that affect benign fitting by jointly analyzing model stability and optimization error.}\label{fig:key_idea}
\end{figure*}

\textbf{Contributions.} 
To this end, this paper introduces \textit{Libra}, \textit{a joint model stability and convergence analysis framework for algorithm-dependent excess risk minimization} in FL context. 
Libra bounds the \textit{minimum excess risk dynamics} (see Definition~\ref{def:minimum_erd}), capturing minimum excess risk of a model iteration trajectory, and characterizes the optimal balance between model stability and gradient norms (see Corollary~\ref{corollary:excess_dynamics} and Theorem~\ref{theorem:joint_excess_risk}). Theoretical results contribute insights into training mechanisms from a novel perspective.

In detail, we formulate Libra toward the general federated optimization in Algorithm~\ref{alg:fedopt} under non-convex regimes, revealing new insights of the three model statuses: under-fitting, benign-fitting, and over-fitting as shown in Figure~\ref{fig:key_idea}. 
We use Libra to predict how algorithm-dependent hyperparameters (e.g., global learning rate, server momentum, and local update steps) influence model generalization in FL. 
Concretely, we found a strong relation between the global learning rate selection and model benign-fitting status in Theorem~\ref{theorem:joint_excess_risk}. 
Besides, we demonstrate that large local update steps and server momentum worsen the model stability, indicating a large generalization gap. Surprisingly, they yield a lower minimum excess risk dynamics by better trading off accelerating convergence and worsening stability in Theorem~\ref{theorem:excess_fosm}. 
Empirically, we evaluate our theories in standard federated learning settings. Although our results are built on the context of FL, our insights can be extended to general distributed optimization. 
Moreover, this work is a novel extension of model stability theories~\citep{lei2020fine, lei2023stability}, additionally considering gradient norms dynamics for quantifying excess risk in non-convex regimes. And, we provide additional findings on FL generalization~\citep{init2024understanding, sun2024understanding}. Please refer to Appendix A for a comprehensive review of related works.

\section{Problem Formulation}

\renewcommand{\algorithmicrequire}{\textbf{Input:}}
\renewcommand{\algorithmicensure}{\textbf{Output:}}
\begin{algorithm}[t]
\caption{Federated Optimization with Server \\
\textbf{\textit{SGD} (\romannumeral1)} or \textbf{\textit{Momentum} (\romannumeral2)}}
\label{alg:fedopt}
\begin{algorithmic}[1]
\REQUIRE $\bs{x}^0, \eta_l, \eta_g, T,$ $\bs{m}^0$,$0<\nu$, $0\leq\beta\leq1$
\ENSURE $\bs{x}^T$
\FOR{round $t$ $\in$ $[T]$}
    \STATE Server broadcast model $\bs{x}^t$ to clients
    \FOR{client $i$ $\in$ $[N]$ in parallel}
        \STATE $\bs{x}^{t,0}_i = \bs{x}^t$
        \FOR{local update step $k = 0,\dots,K-1$ }
            \STATE $\bs{x}^{t,k+1}_i = \bs{x}^{t,k}_i - \eta_l \bs{g}_i^{t,k}$
        \ENDFOR
         \STATE Client uploads local updates $\bs{d}_i^t = \bs{x}^{t,0} - \bs{x}^{t, K}$
    \ENDFOR
    \STATE Server aggregates $\bs{d}^t = \frac{1}{N}\sum_{i\in [N]}\bs{d}_i^t$
    \STATE \textbf{(\romannumeral1)} Server updates $\bs{x}^{t+1} = \bs{x}^t - \eta_g \bs{d}^t$
    \STATE \textbf{(\romannumeral2)} Server computes $\bs{m}^{t} = \beta \bs{m}^{t-1} + \nu \bs{d}^{t}$ and updates $\bs{x}^{t+1} = \bs{x}^t - \eta_g \bs{m}^t$
\ENDFOR
\end{algorithmic}
\end{algorithm}

We consider a standard FL setting with $N$ clients connected to one aggregator. The basic problem in FL is to learn a prediction function parameterized by $\bs{x} \in \mathcal{X} \subseteq \mathbb{R}^d$ to minimize the following global population risk:
\begin{equation} \label{eq:global_obj}
    F(\bs{x}) := \frac{1}{N}{\sum}_{i=1}^N \mathbb{E}_{z \sim \mathcal{P}_i}[\ell(\bs{x}; z)] ,
\end{equation}
where $\mathcal{P}_i$ is the underlying data distribution on client $i$ and may differ across different clients, $\ell(\bs{x}; z):\mathcal{X}\times \mathcal{Z}\rightarrow \mathbb{R}^{+}$ is some nonnegative loss function, $z$ denotes the sample from the sample space $\mathcal{Z}$.
Since $\mathcal{P}_i$ is unknown typically, the global population risk $F(\bs{x})$ cannot be computed. Instead, we can access to a training dataset $\mathcal{D} =\cup_{i=1}^N \mathcal{D}_i$, where $\mathcal{D}_i = \{z_j^{(i)}\}_{j=1}^{n_i}$ is the training dataset on client $i$ with size $n_i$. Then, we can estimate $F(\bs{x})$ using global empirical risk: 
\begin{equation} \label{eq:emp_obj}
    f(\bs{x}) := \frac{1}{N} {\sum}_{i=1}^N f_{i}(\bs{x}) ,
\end{equation}
where $f_{i}(\bs{x}) = \frac{1}{n_i}\sum_{j=1}^{n_i} \ell(\bs{x}; z_j^{(i)})$ is the local empirical risk on client $i$, $z_j^{(i)} \overset{\text{ i.i.d. }}{\sim}\mathcal{P}_i$ represents a sample from local data distribution $\mathcal{P}_i$.

In practice, a family of FL algorithms to solve (\ref{eq:emp_obj}) has been proposed \citep{hsu2019measuring, reddi2020adaptive,xu2021fedcm, zeng2025power, wang2020tackling}, which take the form of alternating optimization between clients and server:
\begin{align*}
    \text{\textbf{Client:}}& \quad \bs{x}_i^{t,k+1} = \textsc{ClientOpt}(\bs{x}_i^{t,k}, \bs{g}_i^{t,k}, \eta_l); \\ 
    \text{\textbf{Server:}}& \quad \bs{x}^{t+1} = \textsc{ServerOpt}(\bs{x}^{t}, \bs{d}^{t}, \eta_g),
\end{align*}
where \textsc{ClientOpt} and \textsc{ServerOpt} are gradient-based optimizers with learning rates $\eta_l$ and $\eta_g$ respectively. Essentially, \textsc{ClientOpt} aims to minimize the loss on each client's local data with stochastic gradient $\bs{g}_i^{t,k}$ computed on the $k$-th local SGD steps on client $i$'s local model $\bs{x}_i^{t,k}$ at communication round $t$. Then, \textsc{ServerOpt} updates the global model based on the aggregated clients' model updates $\bs{d}^{t}$ (see Line 8 in Algorithm~\ref{alg:fedopt}). 
In this work, we consider two particular cases where both \textsc{ClientOpt} and \textsc{ServerOpt} are in the form of SGD as shown in Algorithm~\ref{alg:fedopt}. Specifically, we present the details of Libra in standard \textbf{federated optimization with server SGD (\romannumeral1)}, which notably recovers FedAvg~\citep{mcmahan2017communication} when $\eta_g=1$. 
Then, we extend Libra to \textbf{federated optimization with server momentum (\romannumeral2)} (FOSM) to show that Libra is general for characterizing how server momentum improves generalization. Notably, FOSM is a foundamental form of FedAvgM~\citep{hsu2019measuring} ($\nu = 1$) and FedCM~\citep{xu2021fedcm} ($\nu = 1-\beta$). Further extension of FOSM induces FedAdam~\citep{reddi2020adaptive}, FedAMS~\citep{wang2022communication}, and FedGM~\citep{sun2024role}. Hence, understanding the momentum mechanism from a generalization perspective is promising.


\textbf{Excess risk.}
Analyzing the generalization of training mechanisms is vital for designing generalizable algorithms~\citep{lei2020fine, teng2021towards, init2024understanding}. We focus on the \textit{excess risk}\footnote{Some works~\citep{teng2021towards} define \textit{excess risk} as $\mathcal{E}(\bs{x})=F(\bs{x})- F(\bs{x}^*)$, where $\bs{x}^*:=\min_{\bs{x}\in \mathcal{X}}F(\bs{x})$ denotes the population risk minimizer. Then, they decompose it into $\mathcal{E}(\bs{x})=\mathcal{E}_G+ \mathcal{E}_O + f(\hat{\bs{x}})-F(\bs{x}^*)$. Despite the differences in definition, one still needs to bound optimization error and generalization gap.} $\mathcal{E}(\bs{x}):=F(\bs{x})- f(\hat{\bs{x}})$, where $\hat{\bs{x}}$ denotes the empirical risk minimizer. It can be decomposed as
\begin{equation}\label{eq:excess_risk}
    \mathcal{E}(\bs{x}) = \underbrace{F(\bs{x})-f(\bs{x})}_{\mathcal{E}_G: \text { generalization gap}}+\underbrace{f(\bs{x})-f(\hat{\bs{x}})}_{\mathcal{E}_O: \text { optimization error }},
\end{equation}
where $\mathcal{E}_G$ denotes the estimation error due to the approximation of the unknown data distribution $\mathcal{P}$, and $\mathcal{E}_O$ represents the optimization error on training data $\mathcal{D}$. 

\section{Understanding the Tightrope between \\ Model Stability and Gradient Norms}

This section draws important insights in federated optimization and elaborates on the details of the Libra framework in the context of Algorithm~\ref{alg:fedopt}. 
In general, Libra consists of three main steps: (\romannumeral1) Deriving a general upper bound of model stability dynamics; (\romannumeral2) Deriving a general upper bound of gradient norm dynamics; (\romannumeral3) Assembling the upper bound of excess risk dynamics according to Corollary~\ref{corollary:excess_dynamics}, and optimizing the assembled upper bound w.r.t algorithm-dependent hyperparameters.

\subsection{Establishing Excess Risk Dynamics}\label{sec:libra1}

\textit{Algorithmic stability} is a promising theoretical framework for analyzing the generalization gap~\citep{hardt2016train, lei2020fine}. In which, \textit{on-average model stability}~\citep{lei2020fine} gives fine-grained analysis that yields tighter bounds than \textit{uniform stability}~\citep{hardt2016train}. 
In detail, model stability means that any perturbation of samples on datasets cannot lead to a big change in the model parameters trained by learning algorithms in expectation. 
Formally, model stability derives an upper bound of the generalization gap $\mathcal{E}_G(\bs{x})$ about model parameters. For instance, we refer to Theorem 2, (b)~\citep{lei2020fine}:
\begin{theorem}[Generalization gap via $\ell_2$ model stability]\label{thm:gen} Let $\mathcal{D}, \mathcal{D}^\prime$ differ in at most one data sample (perturbed sample). For the two models $\bs{x}$ and $\tilde{\bs{x}}$ trained by a learning algorithm on these two sets, if function $f$ is $L$-smooth, the generalization gap is upper bounded:
\begin{equation}\label{eq:gen_bound}
\begin{aligned}
\mathcal{E}_G(\bs{x}) \leq \frac{L+\gamma}{2} \mathbb{E}_{\mathcal{D}, \mathcal{D}^\prime}[\|\bs{x}-\tilde{\bs{x}}\|^2] + \frac{1}{2\gamma} \mathbb{E}\|\nabla f(\bs{x})\|^2,
\end{aligned}
\end{equation}
where $\gamma>0$ is a free parameter for tuning depending on the properties of the objectives. 
\end{theorem}

\begin{remark}
To unify the stability analysis with convergence analysis, we rigorously modified the original empirical loss into the squared gradient norms, as justified in Appendix B.
Early stability analysis~\citep{hardt2016train, lei2020fine} typically considered that stability dominates the generalization gap $\mathcal{E}_G$, while $\mathcal{E}_O$ is minor if the model is fully-trained. Hence, stability theory implicitly proves that if an algorithm is stable, its excess risk is small. 
In this work, we highlight Theorem~\ref{thm:gen} to emphasize that the generalization gap is subject to the gradient norm. As the gradient norm $\mathbb{E}\|\nabla f(\bs{x})\|^2$ typically is non-zero in neural network training, omitting the effects of gradient norms may lead to a suboptimal generalization gap.
\end{remark}

In practice, we focus more on how the neural networks' test loss evolves during training, which indicates their generalization performance. Hence, we turn to analysis \textit{excess risk dynamics} for directly quantifying test loss evolution:
\begin{corollary}[Excess risk dynamics]\label{corollary:excess_dynamics} 
Under conditions of Theorem~\ref{thm:gen}, a training algorithm creates a model trajectory $\{\bs{x}^t\}_{t=0}^{T}$ on dataset $\mathcal{D}$, and a virtual trajectory $\{\tilde{\bs{x}}^t\}_{t=0}^{T}$ on datasets $\mathcal{D}^\prime$. Substituting \eqref{eq:gen_bound} into \eqref{eq:excess_risk} and using $\mathcal{E}_{O}(\bs{x}^t) \leq C\cdot\mathbb{E}\|\nabla f(\bs{x}^t)\|^2$ for some $C>0$, the any-time excess risk is bounded by a weighted sum of model stability and gradient norm:
$$
\begin{aligned}
\mathcal{E}(\bs{x}^t) \leq \frac{L+\gamma}{2} \underbrace{\mathbb{E}\|\bs{x}^t - \tilde{\bs{x}}^t\|^2}_{\text{Model stability}} + (\frac{1}{2\gamma} + C)\underbrace{\mathbb{E}\|\nabla f(\bs{x}^t)\|^2}_{\text{Gradient norm}}.
\end{aligned}
$$
\end{corollary}

\begin{remark}[Interpretation of $C$]
We note that the value of $C$ quantifies the relation between gradient norms and the descent quantity of the empirical loss. 
For example, if we assume the objective $f$ satisfies the $\mu$-Polyak-Lojasiewicz (PL) condition of the typical assumption in non-convex functions~\citep{jain2017non}, it yields $C = 1/2\mu$. However, it is noted that the PL condition is sometimes considered too strong for non-convex optimization analysis. The Kurdyka-Lojasiewicz (KL) condition is a milder assumption which yields $\mathcal{E}_{O}(\bs{x}^t) \leq C^\frac{1}{2\theta} \cdot\mathbb{E}\|\nabla f(\bs{x}^t)\|^\frac{1}{\theta}$ with smooth $f$ and some $\theta \in (0, 1]$. And, $\theta = 1/2$ recovers the PL condition. Compared with Theorem~\ref{thm:gen}, term $C$ emphasizes the importance of considering gradient norms in excess risk dynamics, which cannot be relaxed by tuning $\gamma$. And, the key idea of Libra it to use the gradient norm $\mathbb{E}\|\nabla f(\bs{x})\|^2$ to bound the optimization error $\mathcal{E}_{O}(\bs{x})$ for further excess risk analysis. 
To maintain generality, we do not consider how the KL condition $\theta$ improves the convergence~\citep{karimi2016linear} and generalization~\citep{charles2018stability} to avoid strong assumptions.
\end{remark}

\begin{remark}[Conflicts in excess risk dynamics]
The model stability and gradient norm dynamics conflict in the excess risk dynamics. For a training process $t\in[T]$, we expect the excess risk to decrease fast and stably, indicating the model's generalization ability keeps improving. 
However, the model stability expands while the gradient norms decreases over model iterations as illustrated in Figure~\ref{fig:key_idea}. In the literature, stability of vanilla SGD typically grows at a speed of $\mathcal{O}(T^{c})$~\citep{hardt2016train, nikolakakis2023beyond, nikolakakis2023select}. And, the convergence speed roughly matches $\mathcal{O}(1/\sqrt{T})$~\citep{reddi2020adaptive}. 
To achieve strong generalization, we hope the stability bounds to grow slowly and the convergence speed to be faster. However, stability and convergence analysis should focus on the same training procedure, while neither of these perspectives can precisely quantify the excess risk alone.
\end{remark}

The conflicts pose two key questions to this scenario: \textbf{\textit{When do gradient norms and model stability reach the optimal balance during training?}} And, \textbf{\textit{How do algorithms' hyperparameters affect their balance?}} Hence, studying the interplay between model stability and gradient norms can be promising, which is not fully discussed to the best of our knowledge. To answer these questions, we aim to bound the \textit{minimum excess risk dynamics} in a model iteration trajectory:
\begin{definition}[Minimum excess risk dynamics]\label{def:minimum_erd}
Given an iterative learning algorithm with proper hyperparameters, it yields a model trajectory $\{\bs{x}\}_{t=0}^{T}$.
The model trajectory achieves the minimum excess risk at a specific time point. We define the excess risk at this particular time as the \textit{minimum excess risk dynamics}, i.e. $\mathcal{E}_{\text{min}} = \min_{t\in[T]}\mathcal{E}(\bs{x}^t)$.
\end{definition}

The minimum excess risk dynamics indicate the best generalization performance of an algorithm may achieve. Understanding the minimum excess risk bound is to identify the algorithm-dependent factors that affect the models' generalization. Hence, it is also crucial to enlighten algorithms design for efficiency and generalization. 

\subsection{Model Stability and Gradient Norms Analysis}\label{sec:libra2}

In this section, we provide the model stability and convergence analysis of Algorithm~\ref{alg:fedopt}, (\romannumeral1) with detailed proof provided in Appendix C. Our analysis relies on standard assumptions on local stochastic gradients~\citep{reddi2020adaptive, init2024understanding, zeng2025power}:
\begin{assumption}\label{asp:unbiasedness}
The local stochastic gradient $\bs{g}_i = \frac{1}{|\xi_i|}\sum_{z\in\xi_i}\nabla \ell(\bs{x}; z)$ (Line 6 in Algorithm~\ref{alg:fedopt}) is unbiased and its variance is $\sigma_l$-bounded, i.e., $\mathbb{E}\left[\bs{g}_i\right] =\nabla f_i(\bs{x})$, $\mathbb{E}\left[\left\|\bs{g}_i-\nabla f_i(\bs{x})\right\|^2\right] \leq \sigma_l^2$, for all $\xi_i \in \mathcal{D}_i, i\in[N]$.
\end{assumption}
\begin{assumption}\label{asp:bgv}
The global variance satisfies $\mathbb{E}\left\|\nabla f_i(\bs{x})-\nabla f(\bs{x})\right\|^2 \leq \sigma_{g}^2$, $\forall i\in[N]$ and $x \in \mathcal{X}$.
\end{assumption}

Then, we prove the FL global model stability:
\begin{theorem}[Global model stability]\label{theorem:global_stability}
Under conditions of Theorem~\ref{thm:gen} and Assumptions~\ref{asp:unbiasedness}\&~\ref{asp:bgv}, if all clients use the same local update step $K$, batch size and $\eta_l = \Theta(\frac{1}{K L})$ for local mini-batch SGD, there is a global learning rate $\eta_g \leq \sqrt{\frac{c}{t}}, c>0$ such that the global model stability is expansive in expectation with $\mathbb{E}\|\bs{x}^{T} - \tilde{\bs{x}}^{T}\|^2 \leq \mathcal{O}\left(K \sigma_n^2 \cdot T^{c\psi}\right)$,
where $\sigma_n^2 = \sigma_l^2+\sigma_g^2/n$ and $\psi = (1 + 4\eta_l L)^{K} \in (1, 2)$.

\textbf{Sketch of proof}\quad
The vanilla algorithmic stability theory analyzes the perturbation of data samples in the training datasets. In FL, the training datasets are decentralized and distributed across clients. Noting that Algorithm~\ref{alg:fedopt} consists of local SGD and global SGD, we consider the global model's stability inherits the local models' stability. Technically, we provide the local stability analysis in Lemma~\ref{lemma:local_stability}. Then, we derive the global model stability by taking the local model stability inherited by the global model during the local updates aggregation steps in FL. 
\end{theorem}

\begin{remark}[Connections with previous works]
Previous FL works~\citep{init2024understanding, sun2024understanding} first extend uniform stability~\citep{hardt2016train} to FL, which mainly analyzes the stability of empirical loss. Despite that, we do not claim contribution to the stability theory, we argue that Theorm~\ref{theorem:global_stability} is the first FL extension of on-average model stability~\citep{lei2020fine, lei2023stability}. 
Compared with vanilla algorithmic stability of SGD~\citep{hardt2016train}, our result matches the rate of $\mathcal{O}(n^{-1})$ w.r.t. the size of the dataset. Beyond the rate of dataset size, the model stability is enlarged by local gradient variance $\sigma_l^2$, and global variance $\sigma_g^2$ from FL settings. 
It means that data heterogeneity across clients worsens the stability of FL algorithms, which matches previous works~\citep{sun2024understanding}. 
Notably, our results first shows that the model stability is linearly scaled by local update steps $K$. This emphasizes that it potentially induces a large generalization gap if the FL system utilizes a large $K$ for communication efficiency~\citep{mcmahan2017communication}. 
\end{remark}

To better clarify the insights of Libra, we provide standard convergence analysis of Algorithm~\ref{alg:fedopt}, (\romannumeral1) for notation consistency and theoretical integrity. Our analysis roughly follows previous non-convex federated optimization works~\citep{li2019convergence, reddi2020adaptive, jhunjhunwala2022fedvarp} with slight modifications for adapting our notations in model stability.
\begin{theorem}[Global gradient norms convergence]\label{theorem:convergence_rate} Under Assumption~\ref{asp:unbiasedness}, ~\ref{asp:bgv}, and $L$-smoothness of $f$, setting $\eta_l \leq \min \{\frac{1}{8KL},\frac{1}{\sqrt{T}KL}\}$, and $\eta_l\eta_g \leq \frac{1}{480KL}$, there exists a $\eta_g$ such that the global model converges to a stationary point at a rate $\min_{t\in[T]}\mathbb{E}\|\nabla f(\bs{x}^t)\|^2 \leq \mathcal{O}\left(\sqrt{\frac{\sigma_K^2 \mathcal{F}}{TK}} + \frac{\sigma_K^2}{T}\right)$, where $\sigma_K^2 = \sigma_l^2 + K\sigma_g^2$, and $f(\bs{x}^0) - f(\hat{\bs{x}}) \leq \mathcal{F}$.
\end{theorem}

Theorem~\ref{theorem:global_stability} shows the expanding speed of model stability, and Theorem~\ref{theorem:convergence_rate} shows the convergence speed of gradient norms. 
Correspondingly, fast convergence of the optimization error (e.g., empirical loss 0 in convex optimization) implements a tighter stability by Theorem~\ref{theorem:global_stability}. Then, \citet{hardt2016train} concludes that \textit{convergence faster, generalization better}, especially in convex regimes. 
However, neural network training is subject to non-zero gradient norms as discussed in Corollary~\ref{corollary:excess_dynamics}; the minimum excess risk bound under such a trade-off has not been fully resolved. \citet{init2024understanding} has derived a similar trade-off observation specifically for FedInit. Notably, it analyzes stability and convergence bounds independently, but regretfully combines the two bounds as the upper bound of excess risk by taking the intersection of hyperparameters (please see Theorem 6~\citep{init2024understanding}). However, its theorem overlooks the trade-off in training dynamics since the stability and convergence rely on the same hyperparameter setup. Besides, their analysis is particularly conducted for FedInit, which can be less meaningful for guiding general FL practice.

\subsection{Libra: Excess Risk Dynamics Minimization}\label{sec:libra3}

This section provides theoretical results of gradient norms and model stability analysis, outputted by the Libra framework. Then, we derive deep insights into minimum excess risk dynamics, regarding algorithm-dependent factors.

Libra is a straightforward but insightful framework. It emphasizes that \textit{model stability and convergence analysis examine the same training process from different perspectives, and the algorithm-dependent hyperparameters should be jointly tuned.} 
In this work, we minimize the upper bound of excess risk dynamics by tuning the global stepsize $\eta_g$ for Corollary~\ref {corollary:excess_dynamics}. And, we derives an \textit{explicit} upper bound of minimum excess risk for Algorithm~\ref{alg:fedopt}: 
\begin{theorem}\label{theorem:joint_excess_risk}\label{theorem:excess_fosm} Aligning with the conditions of Theorem~\ref{theorem:global_stability},~\ref{theorem:convergence_rate}, and letting global learning rate $\eta_g \leq \sqrt{\frac{c}{t}}, c > 0$ and joint learning rate $\eta_l\eta_g \leq \sqrt{\frac{c}{T}}$, \textbf{Algorithm~\ref{alg:fedopt}, option (\romannumeral1)} generates an model trajectory $\{\bs{x}^t\}_{t=0}^{T}$ on the server that yields
\begin{equation}\label{eq:joint_excess_risk}
\begin{aligned}
& \mathcal{E}_{\text{min}} \leq \mathcal{O}\left(\sqrt{\frac{\sigma_K^2 \mathcal{F}}{KT}} + \frac{\sigma_K^2}{T}\right) \\
& \quad + \mathcal{O}\left(\left(\frac{\sigma_n^2 \mathcal{F}^2}{Kc}\right)^{\frac{1}{3}} \cdot \left(\frac{1}{T}\right)^{\frac{1-c\psi}{3}} + \frac{\mathcal{F}}{K\sqrt{Tc}}\right).
\end{aligned}
\end{equation}
And, model trajectory $\{\bs{x}^t\}_{t=0}^{T}$ generated by \textbf{Algorithm~\ref{alg:fedopt}, option (\romannumeral2)} yields
\begin{equation}\label{eq:excess_fosm}
\begin{aligned}
\mathcal{E}_{\text{min}} & \leq \mathcal{O}\left(\sqrt{\beta_{-}\cdot\frac{\sigma_K^2 \mathcal{F}}{KT}} + \beta_{-}\cdot \frac{\sigma_K^2}{T}\right) \\
& \quad + \mathcal{O}\left(\left(\beta_{+}\cdot\frac{\sigma_n^2 \mathcal{F}^2}{Kc}\right)^{\frac{1}{3}} \cdot \left(\frac{1}{T}\right)^{\frac{1-\nu^2c\psi}{3}} + \frac{\mathcal{F}}{K\sqrt{Tc}}\right),
\end{aligned}
\end{equation}
where $\beta_{-} = 1-\beta^T$ and $\beta_{+} = (1+\beta)^T$ denoting additionally attention on how the hyperparameter $\beta$ affects the excess risk. Please see Appendix C, D for proofs accordingly.

\textbf{Sketch of Libra}\quad
We derive that the gradient dynamics $\mathbb{E}\|\nabla f(\bs{x}^t)\|^2$ follows that $\frac{1}{T}\sum_{t=0}^T\mathbb{E}\|\nabla f(\bs{x}^t)\|^2 \leq r_0/ (t \eta_g) + r_1 \eta_g$. Analogously, we also know that $\frac{1}{T}\sum_{t=0}^T\mathbb{E}\| \bs{x}^t - \tilde{\bs{x}}^t\|^2 \leq r_2 \eta_g^2$. 
Libra is required to analyze algorithm-dependent terms $r_0, r_1, r_2$ according to the given learning methods. According to Corollary~\ref{corollary:excess_dynamics}, we roughly have an algorithm-dependent excess risk dynamics $\frac{1}{T}\sum_{t=0}^T\mathcal{E}(\bs{x}^t) \leq r_0/ (t \eta_g) + r_1 \eta_g + r_2 \eta_g^2$. 
Noting that $ \mathcal{E}_{\text{min}} \leq \frac{1}{T}\sum_{t=0}^{T-1} \mathcal{E}(\bs{x}^t)$, minimizing the summarized excess risk dynamics by Lemma~\ref{lemma:constant_stepsize} obtains the final results.
\end{theorem}

\begin{remark}
The theorem proves the upper bound of the minimum excess risk that Algorithm~\ref{alg:fedopt} can achieve with proper hyperparameters. Compared with Theorem~\ref{theorem:convergence_rate}, Equation~\eqref{eq:joint_excess_risk} shows that excess risk always converges more slowly than gradient norms due to model stability. Besides, Equation~\eqref{eq:excess_fosm} demonstrates how momentum $\beta$ affects generalization. 
\end{remark}

Our insights emphasize that algorithm-dependent hyperparameters should be set to minimize excess risk dynamics instead of optimization error, especially in modern neural network training. We now discuss three model statuses: under-fitting, benign-fitting, and over-fitting in Figure~\ref{fig:key_idea}.

\textbf{Global learning rate matters to FL generalization.}\quad
The selection of $\eta_g$ coordinates the model stability and gradient norm dynamics in the excess risk. 
If we expect the excess risk to decrease stably during the $T^*$ training rounds, $\eta_g$ should be smaller than $\sqrt{c/t}$ for $t\in[T^*]$. Under the ideal conditions, the training process could achieve the minimum excess risk at an expected time $T^*$ as proved in Theorem~\ref{theorem:joint_excess_risk}. 
We say it is benign-fitting if the minimum excess risk is obtained at the exact time $T^*$. For all rounds before $T^*$, they are under-fitting as the gradient norm dynamics dominate the excess risk dynamics. Moreover, if the algorithm continues running after time $ T*$, the model stability dynamics dominate the excess risk dynamics. Then, the model overfits afterward. 

\textbf{Large global learning rate makes FL overfit early.}\quad
Given a training procedure with all hyperparameters set except $\eta_g$, the benign-fitting time $T^*$ is highly related to $\eta_g$. At an early stage of the training process, the gradient norm typically decreases faster than the expansion of model stability. Therefore, a relatively large $\eta_g$ is useful for generalization improvement during early training. However, if a large $\eta_g \leq \sqrt{c/\tilde{T}}$ where $\tilde{T} < T^*$, the model starts over-fitting around the early time $\tilde{T}$. 
This is because the condition about $\eta_g$ in Theorem~\ref{theorem:joint_excess_risk} is no longer satisfied for the round after $\tilde{T}$. From the trade-off perspective, a large global learning rate makes the model stability expansive very fast (see Lemma~\ref{lemma:stability_expansion}, Appendix C). 
Therefore, the model stability dominates the excess risk bound earlier with larger $\eta_g$. Then, the excess risk stops decreasing, and the model starts over-fitting. 
In practice, we would like to stop our training procedure as possible as it reaches a near-optimal balance between optimization and generalization error. Hence, our theories can apply to future works in early stopping criteria.

\textbf{Global learning rate decay stabilizes FL generalization.}
Common beliefs in how learning rate decay works come from the optimization analysis of (S)GD~\citep{you2019does, kalimeris2019sgd, ge2019step}: \textit{1) an initially large learning rate accelerates training or helps the network escape spurious local minima; 2) decaying the learning rate helps the network converge to a local minimum and avoid oscillation.} 
Despite the learning rate decay has become common sense in practice, Libra provides another explanation from a new theoretical perspective: \textit{learning rate decay helps the excess risk dynamics decrease stably over a sufficiently long period of the training process}. In detail, a proper learning rate makes Theorem~\ref{theorem:joint_excess_risk} hold after even numerous rounds. 
Supposing an ideal learning rate schedule rule, it can continually balance the relation between gradient norms and model stability, keeping the model asymptotically approaching benign-fitting. 
Moreover, our theories also match recent practice in \textit{Local SGD}~\citep{DBLP:conf/iclr/GuLHA23}, which shows that a small learning rate, long enough training time, and proper local SGD steps improve generalization.

\textbf{Trade-offs on $K$.}\quad
Local update steps $K$ is considered a hyperparameter of Algorithm~\ref{alg:fedopt} or Local SGD, which is a vital trade-off factor. 
For example, the convergence rate suggests clients conduct more steps of local updates for fast convergence. In contrast, large local steps also worsen federated global stability as shown in Theorem~\ref{theorem:global_stability}. Thus, Theorem~\ref{theorem:global_stability} and~\ref{theorem:convergence_rate} show another naive trade-off between model stability and convergence rate on the local update step $K$. 
Theorem~\ref{theorem:joint_excess_risk} shows that setting a learning rate proportional to $1/K$ can resolve the trade-off, yielding that a large $K$ can implement a lower minimal excess risk. Despite that, in an FL system with great system heterogeneity, the local update steps can be unstable due to heterogeneous computation power across clients. In practice, Libra can be a feasible theoretical framework to estimate the best local steps for generalization improvement.

\textbf{FOSM generalizes better with faster gradient norm convergence against enlarged model stability.}\quad
Equation~\eqref{eq:excess_fosm} reveals an additional potential trade-off of FOSM related to the hyperparameter $\beta$. Firstly, the value of $\beta$ determines the convergence rate of the first term. 
Since the first term is slightly slower than the second term, FOSM expected a relatively large $\beta$ to get a tight convergence rate. 
In contrast, large $\beta$ enlarges the last term, denoting worse stability. Thus, $\beta$ also trades off the model stability and optimization error. Besides, many works have observed that SGD with momentum generalizes better than vanilla SGD in model training~\citep{zhou2020towards, jelassi2022towards}. Since FedAvg is commonly viewed as a perturbed version of vanilla SGD~\citep{wang2022unreasonable}, we can provide a new explanation of why FOSM generalizes better than FedAvg from the excess risk bound perspective. 
Please note that if we set $\beta=0$, the FOSM recovers to server SGD (i.e., Equation~\eqref{eq:excess_fosm} recovers to Equation~\eqref{eq:joint_excess_risk}). Since the terms in Equation~\eqref{eq:excess_fosm} hold different rates about $T$ under the same training settings, \textit{there always exists a $\beta \in (0,1)$ that yields a lower minimum excess risk dynamics bound of FOSM}. This indicates that momentum techniques can benefit more from enhancing convergence, regarding worsening model stability. 

\section{Discussions}

This section discusses insights and potential applications, with limitations given in Appendix A.

\textbf{Estimating FL algorithms' generalization ability.}
Following the Libra framework, one can obtain the minimum excess risk dynamics of any iterative gradient-descent-based training algorithm and demonstrate their generalization advantages.
In other words, we argue that the generalization ability of algorithms can be compared with the tightness of $\mathcal{E}_{\text{min}}$ bounds. Besides, we assert that \textit{Learning methods obtain better generalization performance via implementing tighter bounds of $\mathcal{E}_{\text{min}}$ if (\romannumeral1) they either accelerate convergence more than worsening model stability expansion, or (\romannumeral2) stabilize model stability expansion more than slowing convergence.}
For example, we shown that local SGD~\citep{DBLP:conf/iclr/GuLHA23} (related to $K$) or SGD with momentum~\citep{hsu2019measuring}(related to $\beta$) typically generalize better than vanilla SGD, matching the first case. 
For the second case, weight decay has been proved to stabilize model training~\citep{zhou2024towards} and hence improves the model generalization. And, recent work~\citep{zeng2025power} proposes a FL aggregation technique, which may slow convergence while obtaining better generalization.
Hence, systematic analysis and direct comparison $\mathcal{E}_{\text{min}}$ of different algorithms can be a promising future work of Libra. 
Moreover, despite the stability and convergence conflicts that exist in general, developing an optimizer that benefits both aspects is fascinating in the future.

\textbf{Guiding local-global optimization paradigm design.} 
Although this work focuses on assessing Libra within the FL context, its principles could extend to general optimization methods beyond FL. Local SGD or Adam~\citep{cheng2025convergence} have emerged as an efficient way of training large language models (LLMs). For example, ModelSoups~\citep{wortsman2022model} propose to merge the parameters of multiple independently trained LLMs into one model to enhance the generalization performance. Then, DiLoCo~\citep{douillard2023diloco} developed a communication-efficient LLM training framework that periodically merges multiple independently updated models. As both methods align with Algorithm~\ref{alg:fedopt}, we argue that future extensions of Libra may derive practical techniques for enhancing generalization in LLMs' training.

\begin{figure*}[t]
\centering
\subfigure[Using same $\eta_g$, large local update steps $K$ make model over-fitting early.\label{fig:main_theory_a}]{
\begin{minipage}[t]{\textwidth}
\centering
\includegraphics[width=\linewidth]{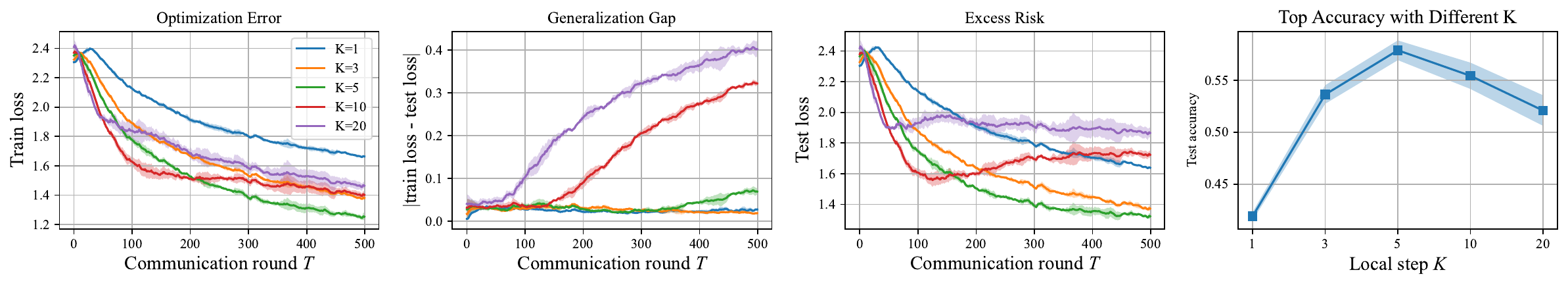}
\end{minipage}}
\\
\subfigure[Stabilizing excess risk dynamics($K=10$) with global learning rate decay $\eta_g^t = \eta_g \cdot \varepsilon^t$.\label{fig:main_theory_b}]{
\begin{minipage}[t]{\textwidth}
\includegraphics[width=\linewidth]{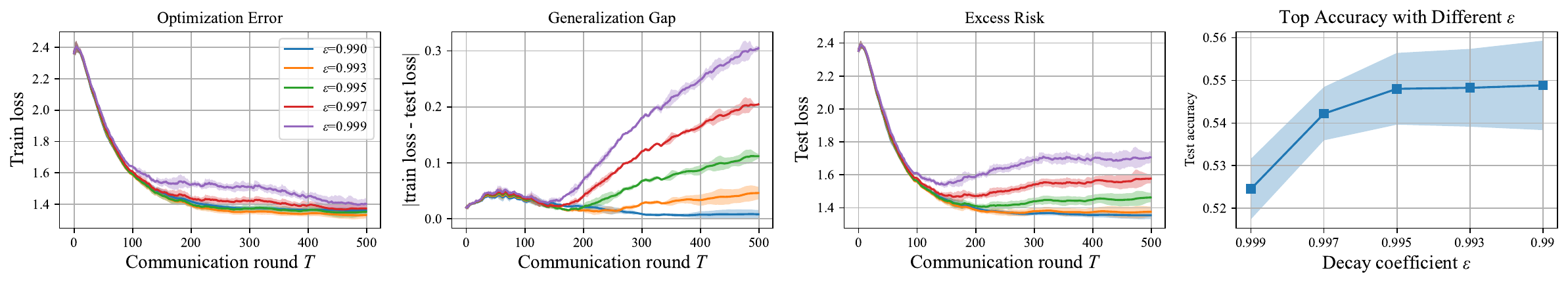}
\end{minipage}}
\\
\subfigure[FOSM: large momentum $\beta$ enlarges generalization gap, indicating worsen stability.\label{fig:main_theory_c}]{
\begin{minipage}[t]{\textwidth}
\includegraphics[width=\linewidth]{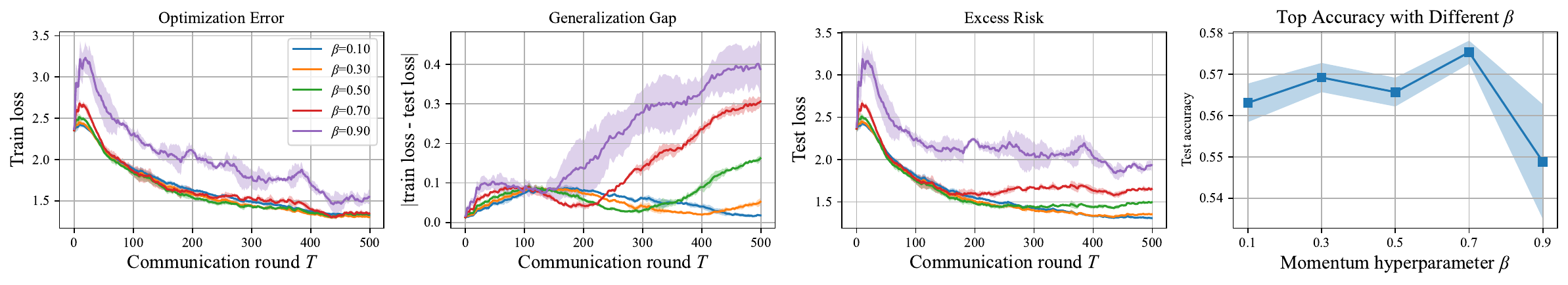}
\end{minipage}}
\caption{Proof-of-concept experiments on Federated CIFAR10 with Dirichlet partition 0.1.}\label{fig:main_theory}
\end{figure*}

\section{Experiments Evaluation}
In this section, we conduct proof-of-concept experiments to validate theory. \textbf{Please note that our experiments are to evaluate our theories instead of pursuing SOTA accuracy, as Algorithm~\ref{alg:fedopt} is a weak baseline in FL.} 
Our experiment setup follows previous works~\citep{mcmahan2017communication, wang2020tackling, zeng2025power, init2024understanding}. We train neural networks on federated partitioned CIFAR-10/100 datasets~\citep{Krizhevsky09learningmultiple}. We generate the datasets of 100 clients following the latent Dirichlet allocation over labels~\citep{hsu2019measuring}, controlled by a coefficient $Dir=0.1$. 
Then, we evaluate our theory with a 4-layer CNN from work~\citep{mcmahan2017communication} for the CIFAR-10 task and ResNet-18-GN~\citep{wu2018group} for the CIFAR-100 task. We select the local learning rate $\eta_l = 0.01$ and global learning rate $\eta_g = 1$ for training hyperparameters. And, we set weight decay as $1e^{-3}$ for the local SGD optimizer. We fix the partial participation ratio of 0.1, the total communication round $T=500$. the local batch size $b = 32$ and local SGD steps $K=\{1,3,5,10,20\}$. 
We reported error bars and conducted t-tests on multiple independent runs to confirm statistical significance. Due to page limitations, CIFAR-10 results are presented in the main paper, while CIFAR-100 results are provided in Appendix E.

\textbf{Model fitting status under different $K$.}
In Figure~\ref{fig:main_theory_a}, we observe that the $K=5$ case is close to benign-fitting at $T=500$ as the test loss converged. And, the $K=1,3$ case is under-fitting before $T=500$. Curve $K=10$ starts over-fitting after 100 rounds, and curve $K=20$ starts over-fitting even earlier. Therefore, when the learning rate is fixed, the local update step should be carefully tuned in practice. Moreover, the training loss curves of $K = \{1, 3, 5\}$ decrease at a rate proportional to $K$, which matches its convergence rate. Meanwhile, the test loss decays at a lower speed than the training loss as the excess risk bound is slower.

\textbf{Generalization gap expands linearly with $K$.} 
Please see the generalization gap plot in Figure~\ref{fig:main_theory_a}. The generalization gap of the global model increases rapidly at a rate proportional to local update steps $K$. For example, the curve $K=20$ reaches the generalization gap value $0.2$ at communication round $150$ while curve $K=10$ reaches the same point around communication round $300$. It is $2\times$ slower due to the choice of $K$ as shown in Theorem~\ref{theorem:global_stability}.


\textbf{Global learning rate decay resolves early over-fitting.} 
The excess risk dynamics grow fast after a short decrease for training processes $K=\{10,20\}$ in Figure~\ref{fig:main_theory_a}. We observe unfavorable convergence behavior and over-fitting. As previously discussed, large local update steps $K$ make the stability bounds in Theorem~\ref{theorem:global_stability} grow too fast. 
Our theories suggest fixing early over-fitting via learning rate decay.
To validate this, we re-run the experiment of $K=10$ in Figure~\ref{fig:main_theory_b} with additional decay of the global learning rate as $\eta_g^t = \eta_g \cdot \varepsilon^t$ for $t\in[T]$. We choose decay coefficient $\varepsilon = \{0.999, 0.997, 0.995, 0.993, 0.99\}$ with the results shown in Figure~\ref{fig:main_theory_b}. The results show that proper global learning rate decay stabilizes the federated optimization.

\textbf{Large $\beta$ enlarge generalization gap.} 
We report the results of FOSM on the CIFAR-10 task with $K=5$ and $b=32$. This empirical evidence for Equation~\eqref{eq:excess_fosm} is shown in Figure~\ref{fig:main_theory_c}. We observe that the training loss curves of FOSM are not very sensitive to low $\beta$. However, the generalization gap is sensitive to the selection of $\beta$. This is because the FOSM model stability expands sub-linearly with $\beta$ (the third term in Equation~\eqref{eq:excess_fosm}) from the generalization gap plots. And, we also observed that FOSM typically implements a better test accuracy than Algorithm~\ref{alg:fedopt}. Despite our theories aligning with the excess risk curves, we clarify that lower excess risk curves do not always correspond to higher test accuracy in practice. In Figure~\ref{fig:main_theory_c}, the best testing accuracy is achieved when $\beta=0.7$, whereas the lowest excess risk occurs when $\beta = 0.1$. Please refer to the calibration studies of classification models~\citep{guo2017calibration} for the explanation.

\section{Conclusion}

This paper proposes the Libra framework for analyzing the minimum excess risk of federated optimization algorithms, which jointly analyzes the model stability and gradient norms trade-offs to characterize the excess risk dynamics precisely. This framework is transferable to arbitrary FL algorithms, as shown in the example analysis on FOSM. Experiments evaluate the theoretical efficacy, demonstrating its potential for guiding FL practice and algorithm design.

\bibliography{references}

\newpage
\onecolumn
\begin{appendices}
In the appendices, we provide additional details and results that are omitted from the main paper.
The appendices are structured as follows:
\begin{itemize}
    \item Appendix A provides comprehensive review of related works and discussions on limitations.
    \item Appendix B provides the proof of Theorem~\ref{thm:gen}.
    \item Appendix C provides the proof of Theorem~\ref{theorem:global_stability}, the proof of Theorem~\ref{theorem:convergence_rate} and Equation~\eqref{eq:joint_excess_risk} in Theorem~\ref{theorem:joint_excess_risk}.
    \item Appendix D provides the proof of Equation~\eqref{eq:excess_fosm} in Theorem~\ref{theorem:joint_excess_risk}.
    \item Appendix E elaborates on the experiment details and missing experiment results.
\end{itemize}
\end{appendices}
\vspace{1em}

\section{A. Related Works \& Limitations}\label{app:related_work}

In this section, we review the related literature on generalization analysis, algorithmic stability, and recent studies on generalization in distributed optimization.

\subsection{Related Works}

\textbf{Generalization analysis.}\quad
Understanding generalization has long been a central objective in learning theory. A classical approach is uniform convergence, which relies on complexity measures such as VC dimension or Rademacher complexity to upper bound the generalization error~\citep{shalev2010learnability, yin2019rademacher}. However, these bounds often fail to capture the practical generalization behavior of modern neural networks, as they depend solely on the hypothesis class and ignore the influence of the training algorithm~\citep{zhang2021understanding}.
An alternative line of work is based on the PAC-Bayes framework, which has been particularly successful in analyzing the generalization of stochastic optimization methods and deep neural networks~\citep{dziugaite2017computing, pensia2018generalization, neu2021information}. Other complementary approaches include information-theoretic bounds\citep{steinke2020reasoning} and compression-based bounds\citep{arora2019fine}, which offer different insights into the generalization behavior of learning algorithms.
It is important to note that our work does not aim to improve the tightness of generalization bounds. Rather, our contributions are orthogonal and complementary to these foundational approaches.

\textbf{Algorithmic stability.}\quad
Algorithmic stability is another powerful tool for explaining generalization. It quantifies how sensitive a learning algorithm is to perturbations in its input dataset. Various notions of stability have been proposed, such as uniform stability\citep{bousquet2002stability}, hypothesis stability\citep{elisseeff2005stability, rubinstein2012stability}, locally elastic stability\citep{deng2021toward}, Bayes stability\citep{li2019generalization}, model stability\citep{liu2017algorithmic, lei2020fine}, gradient stability\citep{lei2023stabilitynonsmooth}, hypothesis set stability\citep{foster2019hypothesis}, and on-average (model) stability\citep{kuzborskij2018data, lei2020fine}.
Stability-based analysis has been widely applied beyond classical settings, including randomized algorithms\citep{elisseeff2005stability}, transfer learning\citep{kuzborskij2018data}, and privacy-preserving learning~\citep{dwork2018privacy}. An important property of the stability analysis is that it considers only the particular model produced by the algorithm, and therefore can use the property of the learning algorithm to imply complexity-independent generalization bounds.

\textbf{Generalization of Local SGD and Federated Learning.}\quad
Federated Averaging (FedAvg)\citep{mcmahan2017communication}, a foundational algorithm in federated learning (FL), was introduced to reduce communication overhead by allowing multiple local updates between global synchronizations. Its core idea is closely related to Local SGD, which has been extensively studied in the context of accelerating convergence for large-scale convex and non-convex optimization\citep{lei2023stability}. However, while classical Local SGD research primarily focuses on optimization efficiency, FL introduces additional challenges due to data heterogeneity across clients, which leads to client-drift, inconsistent local optima, and degraded generalization and convergence performance. To address these challenges, numerous studies have investigated the generalization properties of FL algorithms. Many of these works adopt PAC-Bayes or information-theoretic frameworks to derive generalization bounds for the federated setting, often utilizing the Donsker–Varadhan variational representation~\citep{wu2023information, wu2023federated, barnes2022improved}. These theoretical insights have inspired the development of robust FL algorithms aimed at mitigating the negative effects of heterogeneity and improving generalization~\citep{reisizadeh2020robust}. In another line of work, \citet{qu2022generalized} proposes incorporating Sharpness-Aware Minimization (SAM) into FL to encourage flatter loss landscapes, leading to improved generalization. Subsequent efforts~\citep{caldarola2022improving, sun2023dynamic, shi2023make} build on this idea by proposing SAM-based variants tailored for FL. Despite these advances, most existing works focus on the final generalization performance of the global model, without examining the training dynamics that influence generalization throughout the federated optimization process. As a result, there remains a limited understanding of the key algorithmic factors that govern generalization error in FL.

To the best of our knowledge, only two studies provide explicit algorithmic stability analysis in federated settings. \citet{sun2024understanding} extend the stability analysis of~\citet{hardt2016train} to popular FL algorithms including FedAvg~\citep{mcmahan2017communication}, FedProx~\citep{li2020federated}, and SCAFFOLD~\citep{karimireddy2020scaffold}, and demonstrate how data heterogeneity negatively impacts stability and generalization. \citet{init2024understanding} analyzes the stability of their proposed algorithm FedInit, showing that heterogeneity remains a dominant factor in the generalization error. However, their analysis assumes that optimal convergence rates and algorithmic stability can be simultaneously achieved—an assumption that fails in practice, since these objectives often require different hyperparameter settings. This leads to an incomplete treatment of the tradeoff between convergence and stability, potentially misleading model selection and training strategies.
Furthermore, both~\citet{sun2024understanding} and~\citet{init2024understanding} base their analysis on on-average stability with respect to empirical loss~\citep{hardt2016train}. In contrast, our theoretical framework adopts the concept of on-average model stability~\citep{lei2020fine}, which directly reflects the stability of the trained model and allows us to more accurately capture the generalization behavior in FL.

\subsection{Limitations}

\textbf{Limitations in theories.}\quad
This work focuses on assessing Libra within the FL context, though its principles could extend to other optimization methods. We do not propose specific strategies for improving FL algorithms, as the emphasis is on analyzing Libra's theoretical properties. Our theories are built on a simplified FL setting with full client participation and identical local training configurations. Violating this simplification only induces additional variance terms in heterogeneity terms without breaking our key insights~\citep{jhunjhunwala2022fedvarp}. Future research could explore the impact of system heterogeneity on FL generalization and ways to optimize system efficiency in realistic scenarios.

\textbf{Limitations in experiments and discussion.}\quad
This work primarily presents proof-of-concept experiments on image classification tasks using CIFAR datasets, which may raise concerns about the generality and applicability of our findings to broader domains. To provide preliminary evidence of the framework’s applicability beyond vision tasks, we conducted additional experiments in the natural language processing (NLP) domain. Specifically, we fine-tuned a pre-trained TinyBERT model~\citep{jiao2019tinybert} on the AGNews dataset for text classification. For the FL setup, the dataset was partitioned among 100 clients using a Dirichlet distribution with parameter 0.3, comprising 120,000 training samples and 8,000 testing samples. Using local SGD ($\eta_l = 0.1$, $\eta_g = 1$, batch size $b = 64$), we varied the number of local update steps and applied global learning rate decay to observe training dynamics.

Our observations (see anonymous report\footnote{\url{https://github.com/dunzeng/Libra}}) reveal that the generalization gap grows approximately linearly with the number of local update steps, and that a carefully designed global learning rate decay helps stabilize training, both findings consistent with our theoretical analysis.
However, we also note that FOSM does not significantly improve test accuracy in the AGNews task. This can be attributed to the nature of the task, which involves fine-tuning a pre-trained language model; in such settings, accelerating convergence provides limited generalization benefit. We plan to evaluate Libra on large-scale language model pretraining tasks in future work, as discussed in the main paper (see Discussion section).

Furthermore, we acknowledge a key limitation: most modern NLP pipelines rely on Adam or AdamW optimizers~\citep{kunstner2024heavy}, which are structurally different from SGD and less amenable to our current analysis. Extending Libra to accommodate and analyze Adam-like optimizers presents a promising and necessary direction for future research.

\section{B. Generalization by On-average Model Stability}\label{app:model_stability}


Model stability~\citep{lei2020fine} means any perturbation of samples across all clients cannot lead to a big change in the model trained by the algorithm in expectation.
It measures the upper bound of algorithm results (final model) differences trained on similar datasets. And, these similar datasets for evaluating stability are called \textit{Neighbor datasets}~\citep{hardt2016train, sun2024understanding}:
\begin{definition}[Neighbor datasets]\label{def:neighbor_data}
Given a global dataset $\mathcal{D} = \cup_{i=1}^N \mathcal{D}_i = \{z_1, \dots, z_n\}$, where $\mathcal{D}_i$ is the local dataset of the $i$-th client with $|\mathcal{D}_i| = n_i, \forall i\in[N]$. Another global dataset $\tilde{\mathcal{D}} = \cup_{i=1}^N \tilde{\mathcal{D}}_i = \{\tilde{z}_1, \dots, \tilde{z}_n\}$ be drawn independently from $\mathcal{Z}$ such that $z_j, \tilde{z}_{j} \sim \mathcal{P}_i$ if $z_j \in \mathcal{D}_i$.
For any $j = 1, \dots, n$, we define $\mathcal{D}^{(j)} = \{z_1,\dots, z_{j-1}, \tilde{z}_j, z_{j+1}, \dots, z_n\}$ is neighboring to $\mathcal{D}$ by replacing the $j$-th element with $\tilde{z}_j$.
\end{definition}
Then, given a learning algorithm $\mathcal{A}(\cdot)$ and a training dataset $\mathcal{D}$, we denote $\bs{x} = \mathcal{A}(\mathcal{D})$ as the model generated by the method $\mathcal{A}$ with given data. Then, the upper bound of the model's generalization gap is established in the following theorem:

\begin{theorem}[Modified generalization via model stability] Let $\mathcal{D}, \tilde{\mathcal{D}}, \mathcal{D}^{(j)}$ be constructed as Definition~\ref{def:neighbor_data}.  
Let $\gamma > 0$, if for any $z$, the function $f(\bs{x}; z)$ is nonnegative and $L$-smooth, then 
\begin{equation}
\begin{aligned}
& \mathbb{E}_{\mathcal{D}, \mathcal{A}}[F(\mathcal{A}(\mathcal{D}))- f(\mathcal{A}(\mathcal{D}))] \leq \frac{1}{2\gamma} \mathbb{E}_{\mathcal{D},\mathcal{A}}\|\nabla f(\mathcal{A}(\mathcal{D}))\|^2 + \frac{L+\gamma}{2} \frac{1}{n}\sum_{j=1}^n \mathbb{E}_{\mathcal{D}, \tilde{\mathcal{D}}, \mathcal{A}}[\|\mathcal{A}(\mathcal{D}^{(j)}) - \mathcal{A}(\mathcal{D})\|^2].
\end{aligned}
\end{equation}
\end{theorem}



Recalling Definition~\ref{def:neighbor_data}, $\mathcal{D}^{(j)}$ is the neighbor datasets of $\mathcal{D}$ by replacing the $j$-th element $z_j$ with $\tilde{z}_j$, given the definition of the generalization error, we know
$$
\begin{aligned}
&\quad  \mathbb{E}_{\mathcal{D}, \mathcal{A}}[F(\mathcal{A}(\mathcal{D}))- f(\mathcal{A}(\mathcal{D}))] \\
  & =  \mathbb{E}_{\mathcal{D}, \mathcal{A}}\Big[\frac{1}{n}\sum_{j=1}^n \big(F(\mathcal{A}(\mathcal{D}))\big)\Big]-   \mathbb{E}_{\mathcal{D}, \mathcal{A}}\Big[f(\mathcal{A}(\mathcal{D}))\Big]\\
   & =  \mathbb{E}_{\mathcal{D}, \mathcal{A}}\Big[\frac{1}{n}\sum_{j=1}^n\mathbb{E}_{\tilde{z}_j}\big[\ell(\mathcal{A}(\mathcal{D}),\tilde{z}_j)\big]\Big]-   \mathbb{E}_{\mathcal{D}, \mathcal{A}}\Big[f(\mathcal{A}(\mathcal{D}))\Big]\\
& =  \mathbb{E}_{\mathcal{D}, \mathcal{A}}\Big[\frac{1}{n}\sum_{j=1}^n\mathbb{E}_{\tilde{z}_j}\big[\ell(\mathcal{A}(\mathcal{D}),\tilde{z}_j)\big]\Big]-   \mathbb{E}_{\mathcal{D}, \mathcal{A}}\Big[\frac{1}{n}\sum_{j=1}^n \ell(\mathcal{A}(\mathcal{D}); z_j)\Big] \quad\quad\quad \triangleright \text{Definition of empirical loss}\\
& =  \mathbb{E}_{\mathcal{D}, \mathcal{A}}\Big[\frac{1}{n}\sum_{j=1}^n\mathbb{E}_{\tilde{\mathcal{D}}}\big[\ell(\mathcal{A}(\mathcal{D}),\tilde{z}_j)\big]\Big]-   \frac{1}{n}\sum_{j=1}^n \mathbb{E}_{\mathcal{D},\tilde{\mathcal{D}},\mathcal{A}}\Big[\ell(\mathcal{A}(\mathcal{D}^{(j)}); \tilde{z}_j)\Big]\\
&=\mathbb{E}_{\mathcal{D},\tilde{\mathcal{D}}, \mathcal{A}}\Big[ \frac{1}{n}\sum_{j=1}^n \big(\ell(\mathcal{A}(\mathcal{D}),\tilde{z}_j)- \ell(\mathcal{A}(\mathcal{D}^{(j)}); \tilde{z}_j)\big)\Big].
    \end{aligned}
$$

Recall that we assume the loss function $\ell(x;z)$ satisfies $L$-smooth, i.e., $\forall \bs{x},\bs{y} \in \mathcal{X}, \forall z \in \mathcal{Z}, \Vert \nabla \ell(\bs{x};z)- \nabla \ell(\bs{y};z) \Vert\leq L\Vert\bs{x}-\bs{y} \Vert $. Noting that the empirical risk $f(\bs{x})=\frac{1}{n}\sum_{j=1}^n \ell(\bs{x};z_j)$  also satisfies $L$-smooth because $\Vert \nabla f(\bs{x})-\nabla f(\bs{y})\Vert =\Vert \frac{1}{n}\sum_{j=1}^n \nabla \ell(\bs{x};z_j) -\frac{1}{n}\sum_{j=1}^n \nabla \ell(\bs{y};z_j)\Vert \leq  \frac{1}{n}\sum_{j=1}^n\Vert \nabla \ell(\bs{x};z_j)-\nabla \ell(\bs{y};z_j)\Vert \leq L\Vert \bs{x}-\bs{y}\Vert$.
Based on this finding, we further have
$$
\begin{aligned}
    \mathbb{E}_{\mathcal{D}, \mathcal{A}}[F(\mathcal{A}(\mathcal{D}))- f(\mathcal{A}(\mathcal{D}))] \leq \mathbb{E}_{\mathcal{D}, \tilde{\mathcal{D}}, \mathcal{A}}\left[\left\langle\mathcal{A}(\mathcal{D}^{(j)})- \mathcal{A}(\mathcal{D}), \frac{1}{n}\sum_{j=1}^n \nabla \ell(\mathcal{A}(\mathcal{D}); \tilde{z}_j)\right\rangle + \frac{L}{2} \left\|\mathcal{A}(\mathcal{D}^{(j)})- \mathcal{A}(\mathcal{D})\right\|^2\right].
\end{aligned}
$$

Using Cauchy's inequality, we have 
$$
\begin{aligned}
\mathbb{E}_{\mathcal{D}, \tilde{\mathcal{D}}, \mathcal{A}}\Big[\langle\mathcal{A}(\mathcal{D}^{(j)})- \mathcal{A}(\mathcal{D}), \frac{1}{n}\sum_{j=1}^n\nabla \ell(\mathcal{A}(\mathcal{D}); \tilde{z}_j)\rangle \Big]  & \leq \mathbb{E}_{\mathcal{D}, \tilde{\mathcal{D}}, \mathcal{A}}\Big[\|\mathcal{A}(\mathcal{D}^{(j)})- \mathcal{A}(\mathcal{D})\| \cdot \|\nabla f(\mathcal{A}(\mathcal{D}))\| \Big]\\
& \leq \frac{\gamma}{2}\mathbb{E}_{\mathcal{D}, \tilde{\mathcal{D}}, \mathcal{A}}\|\mathcal{A}(\mathcal{D}^{(j)})- \mathcal{A}(\mathcal{D})\|^2 + \frac{1}{2\gamma} \mathbb{E}_{\mathcal{D}, \tilde{\mathcal{D}}, \mathcal{A}}\|\nabla f(\mathcal{A}(\mathcal{D}))\|^2.
\end{aligned}
$$

Then, we let the expectation absorb the uniform randomness of samples to simplify the equation:
$$
\mathbb{E}_{\mathcal{D}, \mathcal{A}}[F(\mathcal{A}(\mathcal{D}))- f(\mathcal{A}(\mathcal{D}))] \leq \frac{L+\gamma}{2} \underbrace{\frac{1}{n}\sum_{j=1}^n \mathbb{E}_{\mathcal{D}, \tilde{\mathcal{D}}, \mathcal{A}}[\|\mathcal{A}(\mathcal{D}^{(j)})- \mathcal{A}(\mathcal{D})\|^2]}_{\text{On-average Model stability}} + \frac{1}{2\gamma} \underbrace{\mathbb{E}_{\mathcal{D},\mathcal{A}}\|\nabla f(\mathcal{A}(\mathcal{D}))\|^2}_{\text{Gradient norm}},
$$
where $\gamma > 0$. In the main paper, we simplified some notations such as $\bs{x} = \mathcal{A}(\mathcal{D})$.

\section{C. Libra Framework for Algorithm-dependent Excess Risk Minimization}\label{app:libra}

In this section, we provide technical details of the Libra Framework. 
Libra consists of three main steps:
\begin{itemize}
    \item \textbf{Step 1:} Deriving a general upper bound of model stability dynamics w.r.t algorithm-dependent hyperparameters (e.g., learning rate $\eta_g$).
    \item \textbf{Step 2:} Deriving a general upper bound of model convergence dynamics w.r.t algorithm-dependent hyperparameters.
    \item \textbf{Step 3:} Assembling the upper bound of excess risk dynamics according to Corollary~\ref{corollary:excess_dynamics}, and optimizing the assembled upper bound w.r.t algorithm-dependent hyperparameters.
\end{itemize}  

\subsection{Auxiliary Lemmas}
\begin{lemma}[Tuning the stepsize~\citep{koloskova2020unified}]\label{lemma:constant_stepsize} For any parameters $r_0 \geq 0, b \geq 0, e \geq 0, d \geq 0$ there exists constant stepsize $\eta \leq \frac{1}{d}$ such that
$$
\Psi_T:=\frac{r_0}{\eta T}+b \eta+e \eta^2 \leq 2\left(\frac{b r_0}{T}\right)^{\frac{1}{2}}+2 e^{1 / 3}\left(\frac{r_0}{T}\right)^{\frac{2}{3}}+\frac{d r_0}{T}.       
$$
\end{lemma}

\begin{lemma}[Bounded local drift~\cite{reddi2020adaptive}] \label{lemma:local_drift} Let Assumption~\ref{asp:unbiasedness}~\ref{asp:bgv} hold. When $\eta_l \leq \frac{1}{8KL}$, for any $k\in[K]$, we have 
\begin{equation}\label{eq:local_drift}
    \frac{1}{N}\sum_{i=1}^N \mathbb{E}\left\|\bs{x}_i^{t,k}-\bs{x}^t\right\|^2 \leq 5 K\eta_l^2 (\sigma_l^2 + 6K\sigma_g^2) + 30K^2\eta_l^2\mathbb{E}\left\| \nabla f(\bs{x}^t) \right\|^2
\end{equation}
\end{lemma}

\begin{lemma}\label{axuliary:a}
    For random variables $z_1, \dots, z_n$, we have
\begin{equation}\label{eq:lemma2}
\mathbb{E}\left[\left\|z_1+\ldots+z_n\right\|^2\right] \leq n\mathbb{E}\left[\left\|z_1\right\|^2+\ldots+\left\|z_n\right\|^2\right].
\end{equation}
\end{lemma}

\begin{lemma}\label{axuliary:b}
    For independent, mean 0 random variables $z_1, \dots, z_n$, we have
\begin{equation}\label{eq:lemma3}
\mathbb{E}\left[\left\|z_1+\ldots+z_n\right\|^2\right]=\mathbb{E}\left[\left\|z_1\right\|^2+\ldots+\left\|z_n\right\|^2\right].
\end{equation}
\end{lemma}

\subsection{Step 1: Model Stability Analysis}\label{app:libra_s1}

The vanilla algorithmic stability theory analyzes the perturbation of data samples in the training datasets. In FL, the training datasets are decentralized and distributed across clients. Therefore, the global model's stability inherits the local models' stability. Technically, this section provides the local stability analysis in Lemma~\ref{lemma:local_stability}. 
Then, we derive the global model stability in Lemma~\ref{lemma:stability_expansion} and Theorem~\ref{theorem:global_stability} by taking the stability of local models inherited by the global model by local updates aggregation steps in FL. 

\textbf{Proofs of Local Stability}


\begin{lemma}[Local model stability]\label{lemma:local_stability}
Letting non-negative objectives $f_i, \forall i \in [N]$ satisfies $L$-smooth and Assumptions~\ref{asp:unbiasedness}~\ref{asp:bgv}. We suppose the $i$-th client preserves dataset $\mathcal{D}_i$ and $|\mathcal{D}_i|= n_i$ samples. Its neighbor dataset $\mathcal{D}_i^{(j)}$ has the perturbed sample $\tilde{z}_j \in \mathcal{D}_i$ with probability 1. If the $i$-th client conducts $K$ mini-batch SGD steps with batch-size $b_i$, and non-increasing learning rate $\eta_l = \Theta(\frac{1}{K L})$,  we prove the stability of local iteration on the $i$-th client for $t\in[T]$ as
$$
\mathbb{E}\|\bs{x}_i^{t,K} - \tilde{\bs{x}}_i^{t,K}\|^2 \leq (1 + 4\eta_l L)^{K} \underbrace{\mathbb{E}\|\bs{x}^{t} - \tilde{\bs{x}}^{t}\|^2}_{\text{Global model stability}} + 16 K  (\sigma_l^2 + \frac{3b_i\sigma_g^2}{n_i}) \eta_l^2.
$$
\end{lemma}

\textbf{Proof of Lemma~\ref{lemma:local_stability}}
We investigate the local mini-batch SGD stability on the $i$-th clients with batch size $b_i$. $\mathcal{D}_i$ is the local datasets and $\mathcal{D}_i^\prime$ is the neighbor dataset generated analogous to Definition~\ref{def:neighbor_data}. 
Datasets $\mathcal{D}_i$ and $\mathcal{D}_i^\prime$ differ by at most one data sample.

There are two cases to consider in local training. 

In the first case, local mini-batch SGD selects non-perturbed samples in $\mathcal{D}_i$ and $\mathcal{D}_i^\prime$. Hence, we have
$$
\begin{aligned}
& \quad \mathbb{E}[\|\bs{x}_i^{t,k+1} - \tilde{\bs{x}}_i^{t,k+1}\|^2 | \tilde{z} \notin \xi] \\
& \leq \mathbb{E}\|\bs{x}_i^{t,k} - \tilde{\bs{x}}_i^{t,k} - \eta_l (\bs{g}_i^{t,k} - \tilde{\bs{g}}_i^{t,k}) \|^2 \\
& \leq \mathbb{E}\left[\|\bs{x}_i^{t,k} - \tilde{\bs{x}}_i^{t,k}\|^2 + \eta_l^2 \|\bs{g}_i^{t,k} - \tilde{\bs{g}}_i^{t,k}\|^2 - 2\eta_l \langle\bs{x}_i^{t,k} - \tilde{\bs{x}}_i^{t,k}, \bs{g}_i^{t,k} - \tilde{\bs{g}}_i^{t,k}\rangle\right] \\
& \leq \mathbb{E}\left[\|\bs{x}_i^{t,k} - \tilde{\bs{x}}_i^{t,k}\|^2 + \eta_l^2 \|\bs{g}_i^{t,k} - \tilde{\bs{g}}_i^{t,k}\|^2 - 2\eta_l \langle\bs{x}_i^{t,k} - \tilde{\bs{x}}_i^{t,k}, \nabla f_i(\bs{x}_i^{t,k}) - \nabla f_i(\tilde{\bs{x}}_i^{t,k})\rangle\right] \quad\quad\quad \triangleright \text{Assumption~\ref{asp:unbiasedness}}\\
& \leq \mathbb{E}\left[\|\bs{x}_i^{t,k} - \tilde{\bs{x}}_i^{t,k}\|^2 + \eta_l^2 \|\bs{g}_i^{t,k} - \tilde{\bs{g}}_i^{t,k}\|^2 + 2\eta_l L \|\bs{x}_i^{t,k} - \tilde{\bs{x}}_i^{t,k}\|^2\right] \\
& \leq (1+2\eta_l L)\mathbb{E}\|\bs{x}_i^{t,k} - \tilde{\bs{x}}_i^{t,k}\|^2 + \eta_l^2 \mathbb{E}\|\bs{g}_i^{t,k} - \tilde{\bs{g}}_i^{t,k}\|^2 \\
& = (1+2\eta_l L)\mathbb{E}\|\bs{x}_i^{t,k} - \tilde{\bs{x}}_i^{t,k}\|^2 + \eta_l^2 \mathbb{E}\|\bs{g}_i^{t,k} \pm \nabla f_i(\bs{x}_i^{t,k}) - \tilde{\bs{g}}_i^{t,k} \pm \nabla f_i(\tilde{\bs{x}}_i^{t,k}) \|^2 \\
& \overset{(\romannumeral1)}{\leq} (1+2\eta_l L) \mathbb{E}\|\bs{x}_i^{t,k} - \tilde{\bs{x}}_i^{t,k}\|^2 + 2\eta_l^2 \mathbb{E}\|\bs{g}_i^{t,k} - \nabla f_i(\bs{x}_i^{t,k}) - \tilde{\bs{g}}_i^{t,k} + \nabla f_i(\tilde{\bs{x}}_i^{t,k}) \|^2 \\
& \quad + 2\eta_l^2 \mathbb{E}\|\nabla f_i(\bs{x}_i^{t,k}) - \nabla f_i(\tilde{\bs{x}}_i^{t,k})\|^2\\
& \leq (1+2\eta_l L)\mathbb{E}\|\bs{x}_i^{t,k} - \tilde{\bs{x}}_i^{t,k}\|^2 + 2\eta_l^2 \mathbb{E}\|\bs{g}_i^{t,k} - \nabla f_i(\bs{x}_i^{t,k})\|^2 + 2\eta_l^2 \mathbb{E}\|\tilde{\bs{g}}_i^{t,k} - \nabla f_i(\tilde{\bs{x}}_i^{t,k})\|^2  \\
& \quad + 2\eta_l^2 \mathbb{E}\|\nabla f_i(\bs{x}_i^{t,k}) - \nabla f_i(\tilde{\bs{x}}_i^{t,k})\|^2 \\ 
& \leq (1+2\eta_l L + 2\eta_l^2L^2)\mathbb{E}\|\bs{x}_i^{t,k} - \tilde{\bs{x}}_i^{t,k}\|^2 + 4 \eta_l^2 \sigma_l^2,
\end{aligned}
$$
where we note that the stochastic gradients $\bs{g}_i^{t,k}$ and $\tilde{\bs{g}}_i^{t,k}$ only differ in model and the mini batch data samples are identical in $(\romannumeral1)$.




In the second case, local mini-batch SGD samples a batch data that involves the perturbed sample $\tilde{z}$ from $\mathcal{D}_i$ and $\mathcal{D}_i^\prime$, which happens with probability $b_i/n_i$. Analogously, we have
$$
\begin{aligned}
& \quad \mathbb{E}[\|\bs{x}_i^{t,k+1} - \tilde{\bs{x}}_i^{t,k+1}\|^2 | \tilde{z} \in \xi] \\
& \leq \mathbb{E}\|\bs{x}_i^{t,k} - \tilde{\bs{x}}_i^{t,k} - \eta_l (\bs{g}_i^{t,k} - \tilde{\bs{g}}_i^{t,k}) \|^2 \\
& \leq \mathbb{E}\left[\|\bs{x}_i^{t,k} - \tilde{\bs{x}}_i^{t,k}\|^2 + \eta_l^2 \|\bs{g}_i^{t,k} - \tilde{\bs{g}}_i^{t,k}\|^2 - 2\eta_l \langle\bs{x}_i^{t,k} - \tilde{\bs{x}}_i^{t,k}, \bs{g}_i^{t,k} - \tilde{\bs{g}}_i^{t,k}\rangle\right] \\
& \leq \mathbb{E}\left[\|\bs{x}_i^{t,k} - \tilde{\bs{x}}_i^{t,k}\|^2 + \eta_l^2 \|\bs{g}_i^{t,k} - \tilde{\bs{g}}_i^{t,k}\|^2 - 2\eta_l \langle\bs{x}_i^{t,k} - \tilde{\bs{x}}_i^{t,k}, \nabla f_i(\bs{x}_i^{t,k}) - \nabla \tilde{f}_i(\tilde{\bs{x}}_i^{t,k})\rangle\right] \\
& \leq \mathbb{E}\Big[\|\bs{x}_i^{t,k} - \tilde{\bs{x}}_i^{t,k}\|^2 + \eta_l^2 \|\bs{g}_i^{t,k} - \tilde{\bs{g}}_i^{t,k}\|^2 \\
& \quad - 2\eta_l \langle\bs{x}_i^{t,k} - \tilde{\bs{x}}_i^{t,k}, \frac{1}{n_i}\sum_{j=1}^{n_i} \left(\nabla \ell (\bs{x}_i^{t,k}; z_j) - \nabla \ell (\tilde{\bs{x}}_i^{t,k}; z_j)\right) + \nabla \ell(\tilde{\bs{x}}_i^{t,k}; z) - \nabla \ell (\tilde{\bs{x}}_i^{t,k}; \tilde{z}))\rangle\Big] \\
& \overset{(\romannumeral1)}{\leq} \mathbb{E}\left[\|\bs{x}_i^{t,k} - \tilde{\bs{x}}_i^{t,k}\|^2 + \eta_l^2 \|\bs{g}_i^{t,k} - \tilde{\bs{g}}_i^{t,k}\|^2 - 2\eta_l \langle\bs{x}_i^{t,k} - \tilde{\bs{x}}_i^{t,k}, \nabla f_i(\bs{x}_i^{t,k}) - \nabla f_i(\tilde{\bs{x}}_i^{t,k})\rangle\right] \\
& \leq \mathbb{E}\left[\|\bs{x}_i^{t,k} - \tilde{\bs{x}}_i^{t,k}\|^2 + \eta_l^2 \|\bs{g}_i^{t,k} - \tilde{\bs{g}}_i^{t,k}\|^2 + 2\eta_l L \|\bs{x}_i^{t,k} - \tilde{\bs{x}}_i^{t,k}\|^2\right] \\
& \leq (1+2\eta_l L)\mathbb{E}\|\bs{x}_i^{t,k} - \tilde{\bs{x}}_i^{t,k}\|^2 + \eta_l^2 \mathbb{E}\|\bs{g}_i^{t,k} - \tilde{\bs{g}}_i^{t,k}\|^2 \\
& \overset{(\romannumeral2)}{\leq} (1+2\eta_l L)\mathbb{E}\|\bs{x}_i^{t,k} - \tilde{\bs{x}}_i^{t,k}\|^2 + \eta_l^2 \mathbb{E}\| \bs{g}_i^{t,k} \pm \nabla f_i(\bs{x}_i^{t,k}) - \tilde{\bs{g}}_i^{t,k} \pm \nabla \tilde{f}_i(\tilde{\bs{x}}_i^{t,k}) \|^2 \\
& \leq (1+2\eta_l L)\mathbb{E}\|\bs{x}_i^{t,k} - \tilde{\bs{x}}_i^{t,k}\|^2 + 4 \eta_l^2 \sigma_l^2 + 2\eta_l^2 \mathbb{E}\|\nabla f_i(\bs{x}_i^{t,k}) - \nabla \tilde{f}_i(\tilde{\bs{x}}_i^{t,k})\|^2 \\
\end{aligned}
$$
where $(\romannumeral1)$ uses the fact that $z,\tilde{z} \overset{i.i.d}{\in} \mathcal{P}_i$ to yield $\mathbb{E}[\nabla \ell(\tilde{\bs{x}}_i^{t,k}; z) - \nabla \ell(\tilde{\bs{x}}_i^{t,k}; \tilde{z})] = 0$ over the randomness of local data distribution. Noting that the perturbed samples $z,\tilde{z}$ are i.i.d., and Assumption~\ref{asp:unbiasedness} holds for SGD with batch size 1 (i.e., $\|\nabla \ell(\bs{x}; z) - \nabla f_i(\bs{x})\| \leq \sigma_l^2$ for all $x\in\mathcal{X}, z\in\mathcal{D}_i \sim \mathcal{P}_i$). 
Hence, denoting virtual local objective $\tilde{f}_i(\tilde{\bs{x}}_i^{t,k}) = \frac{1}{|\mathcal{D}_i^\prime|}\sum_{z\in\mathcal{D}_i^\prime} \ell(\tilde{\bs{x}}_i^{t,k}; z)$, which differs with $f_i(\tilde{\bs{x}}_i^{t,k})$ with a perturbed sample,  $(\romannumeral2)$ uses the case that $\tilde{f}_i(\tilde{\bs{x}}_i^{t,k})$ and $\tilde{\bs{g}}_i^{t,k}$ satisfies Assumption~\ref{asp:unbiasedness}. 
Besides, $\tilde{f}_i(x)$ also satisfies Assumption~\ref{asp:bgv} about the same global objective $f$ with $f_i(x)$ for all $x\in\mathcal{X}$ using $z, \tilde{z} \overset{i.i.d.}{\in} \mathcal{P}_i$ for all $i\in[N]$. Based on the above discussion, we have

\begin{equation}\label{eq:local_drift_bound}
\begin{aligned}
& \quad \mathbb{E}[\|\bs{x}_i^{t,k+1} - \tilde{\bs{x}}_i^{t,k+1}\|^2 | \tilde{z} \in \xi] \\
& \leq (1+2\eta_l L)\mathbb{E}\|\bs{x}_i^{t,k} - \tilde{\bs{x}}_i^{t,k}\|^2 + 4 \eta_l^2 \sigma_l^2 + 2\eta_l^2 \mathbb{E}\|\nabla f_i(\bs{x}_i^{t,k}) - \nabla \tilde{f}_i(\tilde{\bs{x}}_i^{t,k})\|^2 \\
& \leq (1+2\eta_l L)\mathbb{E}\|\bs{x}_i^{t,k} - \tilde{\bs{x}}_i^{t,k}\|^2 + 4 \eta_l^2 \sigma_l^2 + 2\eta_l^2 \mathbb{E}\|\nabla f_i(\bs{x}_i^{t,k}) \pm \nabla f(\bs{x}_i^{t,k}) - \nabla \tilde{f}_i(\tilde{\bs{x}}_i^{t,k}) \pm \nabla f(\tilde{\bs{x}}_i^{t,k}) \|^2 \\
& = (1+2\eta_l L)\mathbb{E}\|\bs{x}_i^{t,k} - \tilde{\bs{x}}_i^{t,k}\|^2 + 4 \eta_l^2 \sigma_l^2 \\
&\quad + 2 \eta_l^2 \mathbb{E}\|(\nabla f_i(\bs{x}_i^{t,k}) - \nabla f(\bs{x}_i^{t,k})) - (\nabla \tilde{f}_i(\tilde{\bs{x}}_i^{t,k}) - \nabla f(\tilde{\bs{x}}_i^{t,k})) + (\nabla f(\bs{x}_i^{t,k})) - \nabla f(\tilde{\bs{x}}_i^{t,k})\|^2 \\
& \leq (1+2\eta_l L)\mathbb{E}\|\bs{x}_i^{t,k} - \tilde{\bs{x}}_i^{t,k}\|^2 + 4 \eta_l^2 \sigma_l^2 + 12 \eta_l^2 \sigma_g^2 + 6\eta_l^2 \mathbb{E}\|\bs{x}_i^{t,k} - \tilde{\bs{x}}_i^{t,k}\|^2 
\quad\quad\quad \triangleright \text{Assumption~\ref{asp:bgv}} \\
& \leq (1+2\eta_l L+6\eta_l^2 L^2)\mathbb{E}\|\bs{x}_i^{t,k} - \tilde{\bs{x}}_i^{t,k}\|^2 + 4 \eta_l^2 (\sigma_l^2 + 3\sigma_g^2),
\end{aligned}
\end{equation}

\textbf{Bounding local model stability.} Now, we can combine these two cases. Our bound relies on the probability of whether the perturbed samples are involved. Thus, we have:
$$
\begin{aligned}
& \quad \mathbb{E}_{\mathcal{D}_i,\mathcal{D}_i^\prime}\|\bs{x}_i^{t,k+1} - \tilde{\bs{x}}_i^{t,k+1}\|^2  \\
& \leq P(\tilde{z} \notin \xi) \cdot \mathbb{E}[\|\bs{x}_i^{t,k+1} - \tilde{\bs{x}}_i^{t,k+1}\|^2 | \tilde{z} \notin \xi] + P(\tilde{z} \in \xi) \cdot \mathbb{E}[\|\bs{x}_i^{t,k+1} - \tilde{\bs{x}}_i^{t,k+1}\|^2 | \tilde{z} \in \xi] \\
& \leq (1-\frac{b_i}{n_i})\big((1+2\eta_l L + 2\eta_l^2L^2)\mathbb{E}\|\bs{x}_i^{t,k} - \tilde{\bs{x}}_i^{t,k}\|^2 + 4 \eta_l^2 \sigma_l^2\big) \\
&\quad + \frac{b_i}{n_i}\big((1+2\eta_l L+6\eta_l^2 L^2)\mathbb{E}\|\bs{x}_i^{t,k} - \tilde{\bs{x}}_i^{t,k}\|^2 + 4 \eta_l^2 (\sigma_l^2 + 3\sigma_g^2)\big) \\
& = (1 + 2\eta_l L + 2(1+2b_i/n_i)\eta_l^2 L^2) \mathbb{E}\|\bs{x}_i^{t,k} - \tilde{\bs{x}}_i^{t,k}\|^2 + 4 (\sigma_l^2 + \frac{3b_i\sigma_g^2}{n_i}) \eta_l^2 \quad\quad\quad \triangleright \text{$b_i/n_i \leq 1/2$} \\
& \leq (1 + 2\eta_l L + 4\eta_l^2 L^2) \mathbb{E}\|\bs{x}_i^{t,k} - \tilde{\bs{x}}_i^{t,k}\|^2 + 4 (\sigma_l^2 + \frac{3b_i\sigma_g^2}{n_i}) \eta_l^2 \quad\quad\quad \triangleright \text{$1+2\eta_lL \leq 2$} \\
& \leq (1 + 4\eta_l L) \mathbb{E}\|\bs{x}_i^{t,k} - \tilde{\bs{x}}_i^{t,k}\|^2 + 4 (\sigma_l^2 + \frac{3b_i\sigma_g^2}{n_i}) \eta_l^2, \\
\end{aligned}
$$
where the first equation uses $\xi$ to denote a mini-batch data sampled from the local dataset.

According to the FL protocol, we know $\bs{x}^t = \bs{x}^{t,0}, \tilde{\bs{x}}^t = \tilde{\bs{x}}^{t,0}$. Then, we unroll the above equation over local steps from $K-1$ down to $k=0$ ($K > 1$):
$$
\begin{aligned}
\mathbb{E}_{\mathcal{D}_i,\mathcal{D}_i^\prime}\|\bs{x}_i^{t,K} - \tilde{\bs{x}}_i^{t,K}\|^2 
& \leq (1 + 4\eta_l L)^{K} \mathbb{E}\|\bs{x}^{t} - \tilde{\bs{x}}^{t}\|^2 +  \frac{(1 + 4\eta_l L)^{K}-1}{4\eta_l L} \cdot 4\eta_l^2  (\sigma_l^2 + \frac{3b_i\sigma_g^2}{n_i}).
\end{aligned}
$$


Noting that the conditions on local learning rate $\frac{1}{16KL} \leq \eta_l \leq \frac{1}{8KL}$ from Lemma~\ref{lemma:local_drift} and \ref{lemma:bounded_divergence}, it yields $(1 + 4\eta_l L)^{K} \leq 2$ and $\frac{1}{\eta_l L} \leq 16K$. We have
\begin{equation}\label{app:eq_local_datability}
\mathbb{E}_{\mathcal{D}_i,\mathcal{D}_i^\prime}\|\bs{x}_i^{t,K} - \tilde{\bs{x}}_i^{t,K}\|^2 \leq (1 + 4\eta_l L)^{K} \mathbb{E}\|\bs{x}^{t} - \tilde{\bs{x}}^{t}\|^2 + 16 K  (\sigma_l^2 + \frac{3b_i\sigma_g^2}{n_i}) \eta_l^2.
\end{equation}

\textbf{Proofs of Global Stability}

After the local updates are used to update the global model, the global model inherits the expansive property:

\begin{lemma}[Expansion of global model stability]\label{lemma:stability_expansion} Under conditions of Corollary~\ref{corollary:excess_dynamics} and Assumptions~\ref{asp:unbiasedness}~\ref{asp:bgv}, if all clients use identical local update step $K$, batch size $b$ and $\eta_l = \Theta(\frac{1}{K L})$ for local mini-batch SGD, we denote $\{\bs{x}^t\}_{t=0}^{T}$ is the main trajectory of global model generated by the Algorithm~\ref{alg:fedopt} on dataset $\mathcal{D}$. And, $\{\tilde{\bs{x}}^t\}_{t=0}^{T}$ is the virtual trajectory of the global model trained on datasets $\mathcal{D}^\prime$. The global model stability is expansive in expectation as:
\begin{equation}\label{eq:stability_expansion}
\begin{aligned}
\mathbb{E}\|\bs{x}^{t+1} - \tilde{\bs{x}}^{t+1}\|^2 \leq \left((1-\eta_g)^2 + \eta_g^2 (1 + 4\eta_l L)^{K}\right) \mathbb{E}\|\bs{x}^{t} - \tilde{\bs{x}}^{t}\|^2 + 16 K (\sigma_l^2 + \frac{3b\sigma_g^2}{n}) \eta_l^2 \eta_g^2.
\end{aligned}
\end{equation}
\end{lemma}

By definition of the update rule, we know
$$
\begin{aligned}
\mathbb{E}\|\bs{x}^{t+1} - \tilde{\bs{x}}^{t+1}\|^2 & = \mathbb{E}\|\bs{x}^{t} - \eta_g\bs{d}^t - \tilde{\bs{x}}^{t} + \eta_g\tilde{\bs{d}}^t \|^2 \\
& = \mathbb{E}\|\bs{x}^{t} - \eta_g(\bs{x}^t - \bs{x}^{t,K}) - \tilde{\bs{x}}^{t} + \eta_g(\tilde{\bs{x}}^t - \tilde{\bs{x}}^{t,K}) \|^2 \\
& = \mathbb{E}\|(1-\eta_g)(\bs{x}^{t}-\tilde{\bs{x}}^t) + \eta_g(\bs{x}^{t, K} - \tilde{\bs{x}}^{t,K}) \|^2 \\
& \leq (1+p)(1-\eta_g)^2\mathbb{E}\|\bs{x}^{t}-\tilde{\bs{x}}^t\|^2 + (1+p^{-1})\eta_g^2\mathbb{E}\|\bs{x}^{t, K} - \tilde{\bs{x}}^{t,K}\|^2, \\
\end{aligned}
$$
where $p>0$ is a free parameter.



Lemma~\ref{lemma:local_stability} proves the expectation of local SGD expansion with a perturbed sample. Importantly, the local stability of local updates is conditioned by the uniform stability over the whole FL system. Therefore, we need to consider further the probability $P(\tilde{z} \in \mathcal{D}_i), \forall i \in [N]$. Intuitively, the $i$-th local dataset $\mathcal{D}_i$ and $\tilde{D}_i$ differ with one perturbed sample is proportional to the size of local datasets, i.e., $P(\tilde{z} \in \mathcal{D}_i) = n_i / n, \forall i \in [N]$.  Then, we have the bounded gap of local updates on the $i$-th client considering the uniform model stability:
\begin{equation}\label{eq:global_observed_stab}
\begin{aligned}
& \quad \mathbb{E}\|\bs{x}^{t, K} - \tilde{\bs{x}}^{t,K} \|^2 \\
& = \mathbb{E}\|\frac{1}{N}\sum_{i=1}^N(\bs{x}_i^{t, K} - \tilde{\bs{x}}_i^{t,K})\|^2 \\ 
& \leq \frac{1}{N}\sum_{i=1}^N \left( \mathcal{P}(\tilde{z}\in \mathcal{D}_i) \cdot \mathbb{E}\left[\|\bs{x}_i^{t, K} - \tilde{\bs{x}}_i^{t,K}\|^2 |\tilde{z} \in \mathcal{D}_i\right] + \mathcal{P}(\tilde{z}\notin \mathcal{D}_i) \cdot \mathbb{E}\left[\|\bs{x}_i^{t, K} - \tilde{\bs{x}}_i^{t,K}\|^2 |\tilde{z} \notin \mathcal{D}_i\right] \right) \\
& = \frac{1}{N}\sum_{i=1}^N \left( \frac{n_i}{n} \underbrace{\mathbb{E}\left[\|\bs{x}_i^{t, K} - \tilde{\bs{x}}_i^{t,K}\|^2 |\tilde{z} \in \mathcal{D}_i\right]}_{\eqref{app:eq_local_datability}} + \frac{n-n_i}{n} \underbrace{\mathbb{E}\left[\|\bs{x}_i^{t, K} - \tilde{\bs{x}}_i^{t,K}\|^2 |\tilde{z} \notin \mathcal{D}_i\right]}_{\text{Given below}} \right).
\end{aligned}
\end{equation}
Then, we let $b_i = 0$ for~\eqref{app:eq_local_datability} denote the local stability without the perturbed sample (i.e., $\tilde{z}\notin \mathcal{D}_i$). In this case, the local mini-batch SGD updates are expansive due to local stochastic gradient variance and cumulative SGD steps as:
$$
\mathbb{E}\left[\|\bs{x}_i^{t, K} - \tilde{\bs{x}}_i^{t,K}\|^2 |\tilde{z} \notin \mathcal{D}_i\right] \leq (1 + 4\eta_l L)^{K} \mathbb{E}\|\bs{x}^{t} - \tilde{\bs{x}}^{t}\|^2 + 16 K \eta_l^2 \sigma_l^2.
$$

Combining the above equations, we have
$$
\begin{aligned}
\mathbb{E}\|\bs{x}^{t+1} - \tilde{\bs{x}}^{t+1}\|^2 & \leq (1+p)\left((1-\eta_g)^2 + \eta_g^2 (1 + 4\eta_l L)^{K} \right)\mathbb{E}\|\bs{x}^{t} - \tilde{\bs{x}}^{t}\|^2 \\
& \quad + (1+p^{-1})16 K (\sigma_l^2 + \frac{3b\sigma_g^2}{n}) \eta_l^2 \eta_g^2,
\end{aligned}
$$
where we assume the local SGD of all clients uses the same setting $K=K, b=b_i, \forall i$ for simplicity.

\textbf{Proof of Theorem~\ref{theorem:global_stability}} 

We modify the Equation~\eqref{eq:stability_expansion} with the global learning rate $\eta_g$ subjected to time $t$, and restrict $(1+p)(1-\eta_g)^2 \leq 1$ to have
$$
\mathbb{E}\|\bs{x}^{t+1} - \tilde{\bs{x}}^{t+1}\|^2 \leq \left(1 + \psi (\eta_g^t)^2 \right) \mathbb{E}\|\bs{x}^{t} - \tilde{\bs{x}}^{t}\|^2 + \psi_{\sigma} (\tilde{\eta}^t)^2,
$$
where $\psi = (1+p)(1 + 4\eta_l L)^{K}$, $\psi_{\sigma} = (1+p^{-1})16 K (\sigma_l^2 + \frac{3b\sigma_g^2}{n})$ and $\tilde{\eta}^t = \eta_l\eta_g^t$. Here, we note that $(1+p)(1-\eta_g)^2 \leq 1$ suggests choosing a small global learning rate $\eta_g$.

Then, unrolling the above from time $0$ to $t$ and using that $\bs{x}^{0} = \tilde{\bs{x}}^{0}$, we get
$$
\mathbb{E}\|\bs{x}^{t+1} - \tilde{\bs{x}}^{t+1}\|^2 \leq \psi_{\sigma} \sum_{\tau=0}^t 
 \left(\prod_{k=\tau+1}^t \exp\{\psi \eta_g^{k}\}\right) (\tilde{\eta}^\tau)^2,
$$
where uses the fact that $1+a \leq \exp\{a\}$ for all $a>0$.

Without loss of generality, let $\tilde{\eta}^t \leq \eta_g^t \leq \sqrt{\frac{c}{t}}$ for some $c>0$ to obtain
\begin{equation}
\begin{aligned}
\mathbb{E}\|\bs{x}^{t+1} - \tilde{\bs{x}}^{t+1}\|^2 & \leq \psi_{\sigma} \sum_{\tau=0}^t \left( \prod_{k=\tau+1}^t \exp\{c\psi/k\}\right) \frac{c}{\tau} \\
& = \psi_{\sigma} \sum_{\tau=0}^t \left( \exp\{c\psi \sum_{k=\tau+1}^t {\frac{1}{k}}\}\right) \frac{c}{\tau} \\
& \leq \psi_{\sigma} \sum_{\tau=0}^t \left( \exp\{c\psi \log(\frac{t}{\tau})\}\right) \frac{c}{\tau} \\
& \leq  \psi_{\sigma} t^{c\psi} \sum_{\tau=0}^t  c (\frac{1}{\tau})^{1 + c\psi} \quad\quad\quad \triangleright \text{$\sum_{\tau=0}^t (\frac{1}{\tau})^{1 + c\psi} \leq \frac{1}{c\psi}$}\\
& \leq \frac{\psi_{\sigma}}{\psi} t^{c\psi}.
\end{aligned}
\end{equation}

Noting that $\psi \in (1, 2)$ with $\eta_l = \Theta(\frac{1}{KL})$, we have 
\begin{equation}
\mathbb{E}\|\bs{x}^{T} - \tilde{\bs{x}}^{T}\|^2 \leq \mathcal{O}\left(K (\sigma_l^2+\frac{\sigma_g^2}{n}) \cdot T^{c\psi}\right),
\end{equation}
which concludes the proof. Moreover, we note that tuning the free parameter $p$ can further minimize the stability bound. As this work mainly focuses on deriving insights from minimum excess risk bounds instead of pursuing SOTA tightness of stability bounds, we leave $p$ in $\psi$ here.

\subsection{Step 2: Convergence Analysis}\label{app:libra_s2}

Here, we prove the convergence rate of the global descent rule:
$$
\begin{aligned}
& \text{\textbf{Client: }}\bs{d}_i^{t} = \bs{x}_i^{t, K} - \bs{x}_i^{t, 0} = \eta_l \sum_{k=0}^{K-1} \bs{g}_i^{t, k};\\
& \text{\textbf{Server: }}\bs{x}^{t+1} = \bs{x}^{t}-\eta_g \frac{1}{N}\sum_{i=1}^N \bs{d}_i^{t} = \bs{x}^{t}- \eta_g \boldsymbol{d}^t.
\end{aligned}
$$
Specifically, we analyze the update rule on the server-side gradient descent
$$
\bs{x}^{t+1} = \bs{x}^{t} - \eta_g \bs{d}^t.
$$
Without loss of generality, we rewrite the global descent rule as:
$$
\bs{x}^{t+1} = \bs{x}^{t} - \eta \tilde{\bs{d}}^t = \bs{x}^{t} - \eta \frac{1}{N}\sum_{i=1}^N \frac{\bs{d}_i^t}{K} = \bs{x}^{t} - \frac{\eta}{K} \frac{1}{N} \sum_{i=1}^N \sum_{k=1}^{K}\bs{g}_i^{t,k-1},
$$
where $\eta = K \eta_l\eta_g$ and $\tilde{\bs{d}}^t = \bs{d}^t / \eta_l K$. 

\begin{lemma}[Bounded divergence of regularized global updates]\label{lemma:bounded_divergence} Using $L$-smooth and Assumption~\ref{asp:unbiasedness}~\ref{asp:bgv}, for all client $i\in[N]$ with local iteration steps $k \in [K]$ and local learning rate $\frac{1}{16KL} \leq \eta_l \leq \frac{1}{8KL}$, the differences between uploaded local updates and local first-order gradient can be bounded
\begin{equation}\label{eq:local_divergence}
\mathbb{E}\|\nabla f(\bs{x}^t) - \tilde{\bs{d}}^t\|^2 \leq \frac{15}{16}\mathbb{E}\|\nabla f(\bs{x}^t)\|^2 + 60L^2\left(9\sigma_l^2 + K\sigma_g^2\right)K\eta_l^2.
\end{equation}
\begin{proof}

$$
\begin{aligned}
& \quad \mathbb{E}\|\nabla f(\bs{x}^t) - \tilde{\bs{d}}^t\|^2 \\
& = \mathbb{E}\left\|\frac{1}{N}\sum_{i=1}^N\nabla f_i(\bs{x}^t) - \frac{1}{N}\sum_{i=1}^N \frac{\bs{d}_i^t}{\eta_l K}\right\|^2 \leq \frac{1}{N}\sum_{i=1}^N\mathbb{E}\left\|\nabla f_i(\bs{x}^t) - \frac{\bs{d}_i^t}{\eta_l K}\right\|^2 \\
& = \frac{1}{N}\sum_{i=1}^N \mathbb{E}\left\| \frac{1}{K}\sum_{k=1}^{K}\left(\nabla f_i(\bs{x}^t) \pm \nabla f_i(\bs{x}^{t,k-1}) \right) - \frac{\bs{d}_i^t}{\eta_l K}\right\|^2 \\
& \leq 2\frac{1}{N}\sum_{i=1}^N \frac{1}{K}\sum_{k=1}^{K}\mathbb{E}\|\nabla f_i(\bs{x}^t) - \nabla f_i(\bs{x}^{t,k-1})\|^2 + 2 \mathbb{E}\|\frac{1}{K}\sum_{k=1}^{K} \nabla f_i(\bs{x}^{t,k-1}) - \frac{1}{ K}\sum_{k=1}^{K}\bs{g}_i^{t, k-1}\|^2\\
& \leq 2\frac{1}{N}\sum_{i=1}^N \frac{1}{K}\sum_{k=1}^{K}\mathbb{E}\|\nabla f_i(\bs{x}^t) - \nabla f_i(\bs{x}^{t,k-1})\|^2 + 2 \frac{1}{N}\sum_{i=1}^N\frac{1}{K^2}\sum_{k=1}^{K} \mathbb{E}\| \nabla f_i(\bs{x}^{t,k-1}) - \bs{g}_i^{t, k-1}\|^2\\
\end{aligned} 
$$
Then, using the L-smoothness of $f$ and unbiasedness of local stochastic gradients, we know

$$
\begin{aligned}
\mathbb{E}\|\nabla f(\bs{x}^t) - \tilde{\bs{d}}^t\|^2 & \leq 2L^2 \frac{1}{K}\sum_{k=1}^{K} \frac{1}{N}\sum_{i=1}^N\mathbb{E}\|\bs{x}^t - \bs{x}^{t,k-1}\|^2 + 2\frac{\sigma_l^2}{K} \\
& \leq 2L^2 \left(5 K\eta_l^2 (\sigma_l^2 + 6K\sigma_g^2) + 30K^2\eta_l^2\mathbb{E}\left\| \nabla f(\bs{x}^t) \right\|^2\right)+ 2\frac{\sigma_l^2}{K} \\
& \leq 60K^2L^2\eta_l^2 \mathbb{E}\|\nabla f(\bs{x}^t)\|^2 + (10KL^2 \eta_l^2 + \frac{2}{K})\sigma_l^2 + 60K^2L^2\eta_l^2 \sigma_g^2 \\
& \leq \frac{15}{16}\mathbb{E}\|\nabla f(\bs{x}^t)\|^2 + 10L^2\left((1 + \frac{1}{5\eta_l^2 K^2 L^2})\sigma_l^2 + 6K\sigma_g^2\right)K\eta_l^2, \\
\end{aligned} 
$$
where the last inequality uses $\eta_l \leq \frac{1}{8KL}$ from Lemma~\ref{lemma:local_drift}. Furthermore, we suppose a lower bound of $\eta_l$ guarantees the convergence of the local SGD, which includes $\frac{1}{C K L} \leq \eta_l$. Noting that existing convergence analysis (e.g.,FedVARP~\citep{jhunjhunwala2022fedvarp}) takes $\eta_l \leq \frac{1}{\sqrt{T}KL}$ to minimize the order, it yields that $C > \sqrt{T}$.

In our analysis, we take $\frac{1}{16KL} \leq \eta_l \leq \frac{1}{8KL}$ without loss of generality to obtain
$$
\begin{aligned}
\mathbb{E}\|\nabla f(\bs{x}^t) - \tilde{\bs{d}}^t\|^2 & \leq \frac{15}{16}\mathbb{E}\|\nabla f(\bs{x}^t)\|^2 + 60L^2\left(9\sigma_l^2 + K\sigma_g^2\right)K\eta_l^2.
\end{aligned} 
$$
Moreover, lower $\eta_l$ only induces a larger coefficient on $\sigma_l^2$, which will not break our analysis.
\end{proof}
\end{lemma}

\textbf{Proof of Gradient Norm Convergence (Theorem~\ref{theorem:convergence_rate})}

Using the smoothness of $f$ and taking expectations conditioned on $x^ty$, we have
\begin{equation}\label{eq:descent}
\begin{aligned}
\mathbb{E}\left[f(\bs{x}^{t+1})\right] & = \mathbb{E}[f(\bs{x}^t - \eta \tilde{\bs{d}}^t)] \leq f(\bs{x}^t) - \eta \mathbb{E}[\left \langle \nabla f(\bs{x}^t), \tilde{\bs{d}}^t \right \rangle] + \frac{L}{2}\eta^2 \mathbb{E}\left[\|\tilde{\bs{d}}^t\|^2\right] \\
& \leq f(\bs{x}^t) - \eta \mathbb{E}[\left \langle \nabla f(\bs{x}^t), \tilde{\bs{d}}^t \right \rangle] + \frac{L}{2}\eta^2 \mathbb{E}\left[\|\tilde{\bs{d}}^t\|^2\right] \\
& \leq f(\bs{x}^t) - \eta \mathbb{E}\|\nabla f(\bs{x}^t)\|^2 + \eta \mathbb{E}[\left \langle \nabla f(\bs{x}^t), \nabla f(\bs{x}^t) - \tilde{\bs{d}}^t \right \rangle] + \frac{L}{2}\frac{\eta^2}{\eta_l^2 K^2} \mathbb{E}\left[\|\bs{d}^t\|^2\right] \\
& \leq f(\bs{x}^t) - \frac{\eta}{2} \mathbb{E}\|\nabla f(\bs{x}^t)\|^2 +  \underbrace{\frac{\eta}{2}\mathbb{E}\left[\| \nabla f(\bs{x}^t) - \tilde{\bs{d}}^t\|^2\right]}_{\text{Bounded by}~\eqref{eq:local_divergence}} + \underbrace{\frac{L}{2}\frac{\eta^2}{\eta_l^2 K^2} \mathbb{E}\left[\|\bs{d}^t\|^2\right]}_{\text{Bounded by}~\eqref{eq:local_drift}}, \\
\end{aligned}
\end{equation}
where the last inequality follows since $\langle a,b \rangle \leq \frac{1}{2}\|a\|^2 + \frac{1}{2}\|b\|^2, \forall a,b \in \mathbb{R}^d$.



Substituting corresponding terms of the above equation with \eqref{eq:local_divergence} and \eqref{eq:local_drift}, we have

$$
\begin{aligned}
\mathbb{E}\|\nabla f(\bs{x}^t)\|^2 & \leq  \frac{2(f(\bs{x}^t)-\mathbb{E}\left[f(\bs{x}^{t+1})\right])}{\eta} \\
& \quad +  \frac{15}{16}\mathbb{E}\|\nabla f(\bs{x}^t)\|^2 + 60L^2\left(9\sigma_l^2 + K\sigma_g^2\right)K\eta_l^2 \\
& \quad + L \frac{\eta}{\eta_l^2 K^2} \left(5 K\eta_l^2 (\sigma_l^2 + 6K\sigma_g^2) + 30K^2\eta_l^2\mathbb{E}\left\| \nabla f(\bs{x}^t) \right\|^2 \right).
\end{aligned}
$$

Reorganizing the above equation w.r.t, $\eta_g, \eta_l, K$ and denoting $\tilde{\eta} = \eta_l\eta_g$, we have
\begin{equation}\label{eq:descent_lemma}
\begin{aligned}
\rho \mathbb{E}\|\nabla f(\bs{x}^t)\|^2 & \leq \frac{2(f(\bs{x}^t) - \mathbb{E}\left[f(\bs{x}^{t+1})\right])}{K \tilde{\eta}} + \kappa_1 \tilde{\eta} +\kappa_2 K\eta_l^2,
\end{aligned}
\end{equation}
where
$$
\begin{aligned}
\rho = \frac{1}{16}\cdot(1 - 480LK\tilde{\eta}) > 0, \quad \kappa_1 = 5 L (\sigma_l^2 + 6K\sigma_g^2), \quad \kappa_2 = 60L^2\left(9\sigma_l^2 + K\sigma_g^2\right).
\end{aligned}
$$

Taking full expectation of \eqref{eq:descent_lemma} and summarizing it from time $0$ to $T-1$, we have
\begin{equation}\label{eq:temp_gradient_dynamics}
    \frac{1}{T}\sum_{t=0}^{T-1} \mathbb{E}\|\nabla f(\bs{x}^t)\|^2 \leq \frac{2D}{\rho TK \tilde{\eta}} + \bar{\kappa}_1 \tilde{\eta} + \frac{1}{T}\sum_{t=0}^{T-1} \frac{\kappa_2}{\rho},
\end{equation}
where $\bar{\kappa}_1 = \frac{1}{T}\sum_{t=0}^{T-1} \frac{\kappa_1}{\rho}$ and $f(\bs{x}^0) - f(\hat{\bs{x}}) \leq D$ denotes initialization bias.

Noting that $\rho > 0$ as a constant for sufficiently large $T$, setting $\eta_l \leq \min \{\frac{1}{8KL}$, $\frac{1}{\sqrt{T}KL}\}$, and $\tilde{\eta} \leq \frac{1}{480KL}$,
 there exists an constant $\tilde{\eta}$ such that the convergence rate of Algorithm~\ref{alg:fedopt} w.r.t the global objective $f$ given by:
$$
\min_{t\in[T]}\mathbb{E}\|\nabla f(\bs{x}^t)\|^2 \leq \mathcal{O}\left(\sqrt{\frac{D \sigma_K^2}{TK}}\right) + \mathcal{O}\left(\frac{\sigma_K^2}{T}\right),
$$
where $\sigma_K^2 = (\sigma_l^2 + K\sigma_g^2)$ is variance of local and global gradients.




\subsection{Step 3: Excess Risk Analysis via Joint Minimization}\label{app:libra_s3}


Following Corollary~\ref{corollary:excess_dynamics}, we investigate the upper bound of $\mathcal{E}(\bs{x}^t)$ as follows:
$$
\begin{aligned}
\mathcal{E}(\bs{x}^t) \leq \phi_1 \cdot \mathbb{E}\| \bs{x}^t - \tilde{\bs{x}}^t\|^2 + \phi_2 \cdot \mathbb{E}\|\nabla f(\bs{x}^t)\|^2,
\end{aligned}
$$
where $\phi_1 = \frac{L+\gamma}{2}$ and $\phi_2 = \frac{\gamma+\mu}{2\gamma\mu}$ for notation simplicity.

We propose studying the dynamics of excess risk's upper bound for neural networks' training. In detail, taking a summarization of the above equation from time $t=0$ to $T$, we know
\begin{equation}\label{eq:excess_target}
\begin{aligned}
\min_{t\in[T]} \mathcal{E}(\bs{x}^t) \leq \frac{1}{T}\sum_{t=0}^{T-1} \mathcal{E}(\bs{x}^t) & \leq \phi_1 \cdot \underbrace{\frac{1}{T}\sum_{t=0}^{T-1} \mathbb{E}\| \bs{x}^t - \tilde{\bs{x}}^t\|^2}_{\text{stability dynamics}} + \phi_2 \cdot \underbrace{\frac{1} {T}\sum_{t=0}^{T-1} \mathbb{E}\|\nabla f(\bs{x}^t)\|^2}_{\text{gradient dynamics}}. \\
\end{aligned}
\end{equation}

Now, we assemble stability and gradient dynamics. Then, we jointly minimize the excess risk dynamics upper bound. 

We investigate the general form of stability-bound below
$$
\begin{aligned}
\mathbb{E}\|\bs{x}^{t+1} - \tilde{\bs{x}}^{t+1}\|^2 & \leq \left(1 + \psi (\eta_g^t)^2 \right) \mathbb{E}\|\bs{x}^{t} - \tilde{\bs{x}}^{t}\|^2 + \psi_{\sigma} (\tilde{\eta}^t)^2 \\
& \leq \psi_{\sigma} \sum_{\tau=0}^t 
 \left(\prod_{k=\tau+1}^t \exp\{\psi (\eta_g^{k})^2\}\right) (\tilde{\eta}^\tau)^2 \\
& \leq \psi_{\sigma} \sum_{\tau=0}^t \left( \exp\{c\psi \log(\frac{t}{\tau})\}\right) (\tilde{\eta}^\tau)^2 \\
& \leq \psi_{\sigma} \sum_{\tau=0}^t (\frac{t}{\tau})^{c\psi} (\frac{\tilde{\eta}^\tau}{\tilde{\eta}^t})^2\cdot (\tilde{\eta}^t)^2 \\
& \leq \psi_{\sigma} \sum_{\tau=0}^t (\frac{t}{\tau})^{1+c\psi} (\tilde{\eta}^t)^2 \\
& \leq \frac{\psi_{\sigma}}{c\psi} t^{1+c\psi}  (\tilde{\eta}^t)^2,
\end{aligned}
$$
where $\tilde{\eta}^t \leq \eta_g^t \leq \sqrt{\frac{c}{t}}$ for some $c>0$. Then, we know that global stability is expansive by Theorem~\ref{theorem:global_stability}. Then, substituting $t+1$ with $T$ of the above equation, we have
\begin{equation}\label{eq:temp_stab}
\begin{aligned}
\frac{1}{T}\sum_{t=0}^{T-1} \mathbb{E}\| \bs{x}^t - \tilde{\bs{x}}^t\|^2 & \leq \mathbb{E}\|\bs{x}^{T} - \tilde{\bs{x}}^{T}\|^2 \leq \kappa_{\text{stab}} \cdot  (\tilde{\eta}^t)^2, 
\end{aligned}
\end{equation}
where $\kappa_{\text{stab}} = \frac{\psi_{\sigma}}{c\psi} T^{1+c\psi}$.

\textbf{Joint minimization.} 
Substituting \eqref{eq:excess_target} with stability dynamics in \eqref{eq:temp_stab} and gradient dynamics in \eqref{eq:temp_gradient_dynamics}, we have
$$
\min_{t\in[T]} \mathcal{E}(\bs{x}^t) \leq \underbrace{\frac{\bar{D}}{T \tilde{\eta}} + \bar{\kappa}_1 \cdot \tilde{\eta} + \kappa_{\text{stab}} \cdot (\tilde{\eta})^2}_{\text{stability and convergence trade-off}} + \underbrace{\frac{1}{T}\sum_{t=0}^{T-1} \frac{\phi_1\kappa_2}{\rho}}_{\text{client drift error}},
$$
where
$$
\bar{D} = \frac{2\phi_1 D}{\rho K}, \quad \bar{\kappa}_1 = \frac{1}{T}\sum_{t=0}^{T-1} \frac{\phi_1 \kappa_1}{\rho}, \quad \bar{\kappa}_{\text{stab}} = \frac{\phi_2\psi_{\sigma}}{c\psi} T^{1+c\psi}.
$$
Letting $\tilde{\eta} \leq \sqrt{\frac{c}{T}}$ and Lemma~\ref{lemma:constant_stepsize} yielding $d = \sqrt{\frac{T}{c}}, r_0 = \bar{D}, b = \bar{\kappa}_1$, and $e=\bar{\kappa}_{\text{stab}}$, we obtain
\begin{equation}
\min_{t\in[T]} \mathcal{E}(\bs{x}^t) \leq \left(\frac{\bar{\kappa}_1 \bar{D}}{T}\right)^{\frac{1}{2}}+2 \kappa_{\text{stab}}^{\frac{1}{3}}\left(\frac{\bar{D}}{T}\right)^{\frac{2}{3}}+\frac{\bar{D}}{\sqrt{Tc}} + \frac{1}{T}\sum_{t=0}^{T-1} \frac{\kappa_2}{\rho}.
\end{equation}
Substituting the rate of all terms, we have the final rate of excess risk
\begin{equation}\label{eq:app:constant_excess_risk}
\min_{t\in[T]} \mathcal{E}(\bs{x}^t) \leq \mathcal{O}\left(\sqrt{\frac{\sigma_K^2 D}{KT}}\right) + \mathcal{O}\left(\frac{\sigma_K^2}{T}\right) + \mathcal{O}\left(\left(\frac{\sigma_n^2 D^2}{Kc}\right)^{\frac{1}{3}} \cdot \left(\frac{1}{T}\right)^{\frac{1-c\psi}{3}} + \frac{D}{K\sqrt{Tc}}\right),
\end{equation}
which concludes the proof.

\section{D. Case Study on \textit{FOSM}: Stability and Convergence Trade-off Analysis}\label{app:flmomentum}

\subsection{Model Stability Analysis}

We recall the updated rule of the global model as 
$$
\bs{x}^{t+1} = \bs{x}^{t} - \eta_g \bs{m}^{t}, \quad \bs{m}^{t} = \beta \bs{m}^{t-1} + \nu \bs{d}^{t},
$$
where $\bs{d}^{t} = \frac{1}{N} \sum_{i=1}^N \bs{d}_i^{t}$.
Then, we know
\begin{equation}
\begin{aligned}
& \quad \mathbb{E}\|\bs{x}^{t+1} - \tilde{\bs{x}}^{t+1}\|^2 \\
& = \mathbb{E}\|\bs{x}^{t} - \eta_g\bs{m}^t - \tilde{\bs{x}}^{t} + \eta_g\tilde{\bs{m}}^t \|^2 \\
& = \mathbb{E}\|(\bs{x}^{t} - \tilde{\bs{x}}^{t}) - \eta_g(\bs{m}^t - \tilde{\bs{m}}^t) \|^2 \\
& = \mathbb{E}\|\bs{x}^{t} - \tilde{\bs{x}}^{t} - \eta_g\beta(\bs{m}^{t-1} - \tilde{\bs{m}}^{t-1}) - \eta_g \nu  (\bs{d}^{t} - \tilde{\bs{d}}^{t})\|^2 \quad \quad \quad \triangleright \text{$\bs{m}^{t-1} = (\bs{x}^{t-1} - \bs{x}^t)/\eta_g$}\\
& = \mathbb{E}\|\bs{x}^{t} - \tilde{\bs{x}}^{t} - \beta(-(\bs{x}^t - \tilde{\bs{x}}^t) + (\bs{x}^{t-1} - \tilde{\bs{x}}^{t-1})) - \eta_g \nu  (\bs{d}^{t} - \tilde{\bs{d}}^{t})\|^2 \\
& = \mathbb{E}\|(1+\beta)(\bs{x}^{t} - \tilde{\bs{x}}^{t}) - \beta(\bs{x}^{t-1} - \tilde{\bs{x}}^{t-1}) - \eta_g \nu  (\bs{d}^{t} - \tilde{\bs{d}}^{t})\|^2 \\
& = \mathbb{E}\|(1+\beta)(\bs{x}^{t} - \tilde{\bs{x}}^{t}) - \beta(\bs{x}^{t-1} - \tilde{\bs{x}}^{t-1}) - \eta_g \nu  (\bs{x}^{t} - \bs{x}^{t,K} - \tilde{\bs{x}}^{t} + \tilde{\bs{x}}^{t, K})\|^2 \\
& = \mathbb{E}\|(1+\beta-\eta_g\nu)(\bs{x}^{t} - \tilde{\bs{x}}^{t}) - \beta(\bs{x}^{t-1} - \tilde{\bs{x}}^{t-1}) - \eta_g \nu  (- \bs{x}^{t,K} + \tilde{\bs{x}}^{t, K})\|^2 \\
&  \leq (1+\beta-\eta_g\nu)^2\mathbb{E}\|\bs{x}^{t} - \tilde{\bs{x}}^{t}\|^2 + \beta^2\mathbb{E}\|\bs{x}^{t-1} - \tilde{\bs{x}}^{t-1}\|^2 + \eta_g^2 \nu^2\mathbb{E}\|\bs{x}^{t, K} - \tilde{\bs{x}}^{t, K}\|^2. \\
\end{aligned}
\end{equation}

Substituting the last term of the above equation with \eqref{eq:global_observed_stab}, we obtain
$$
\begin{aligned}
\mathbb{E}\|\bs{x}^{t+1} - \tilde{\bs{x}}^{t+1}\|^2 & \leq (1+\beta-\eta_g\nu)^2\mathbb{E}\|\bs{x}^{t} - \tilde{\bs{x}}^{t}\|^2  + \beta^2\mathbb{E}\|\bs{x}^{t-1} - \tilde{\bs{x}}^{t-1}\|^2 \\
& \quad + \eta_g^2 \nu^2 \left((1 + 4\eta_l L)^{K} \mathbb{E}\|\bs{x}^{t} - \tilde{\bs{x}}^{t}\|^2 + 16 K (\sigma_l^2 + \frac{3b\sigma_g^2}{n}) \eta_l^2 \right)\\
& = \left((1+\beta-\eta_g\nu)^2 + \eta_g^2 \nu^2 \psi \right)\mathbb{E}\|\bs{x}^{t} - \tilde{\bs{x}}^{t}\|^2  + \beta^2\mathbb{E}\|\bs{x}^{t-1} - \tilde{\bs{x}}^{t-1}\|^2 + \nu^2 \psi_{\sigma} (\tilde{\eta}^t)^2 \\
& \leq \left((1+\beta)^2  + \eta_g^2 \nu^2\psi\right)\mathbb{E}\|\bs{x}^{t} - \tilde{\bs{x}}^{t}\|^2  + \beta^2\mathbb{E}\|\bs{x}^{t-1} - \tilde{\bs{x}}^{t-1}\|^2 + \nu^2 \psi_{\sigma} (\tilde{\eta}^t)^2 ,
 \end{aligned}
$$
where the last inequality denotes $\psi = (1 + 4\eta_l L)^{K}$, $\psi_{\sigma} = 16 K (\sigma_l^2 + \frac{3b\sigma_g^2}{n})$ and $\tilde{\eta}^t = \eta_l\eta_g^t$ and modifies $\eta_g$ to $\eta_g^t$. 

Let 
$$
\alpha^t = (1+\beta)^2  + (\eta_g^t)^2 \nu^2\psi, \quad \gamma^t = \nu^2 \psi_{\sigma} (\tilde{\eta}^t)^2
$$
to obtain
$$
\mathbb{E}\|\bs{x}^{t+1} - \tilde{\bs{x}}^{t+1}\|^2 \leq \alpha^t\mathbb{E}\|\bs{x}^{t} - \tilde{\bs{x}}^{t}\|^2  + \beta^2\mathbb{E}\|\bs{x}^{t-1} - \tilde{\bs{x}}^{t-1}\|^2 + \gamma^t
$$

Then, unrolling the recursion from $t$ down to $0$ with $\mathbb{E}\left\|\bs{x}^{0}-\tilde{\bs{x}}^{0}\right\|^2 = 0$, we have the stability of FOSM as
$$
\begin{aligned}
\mathbb{E}\|\bs{x}^{t+1} - \tilde{\bs{x}}^{t+1}\|^2 & \leq \sum_{\tau=0}^t \gamma^\tau \prod_{k=\tau+1}^t (\alpha^k+\beta^2) \\
& = \nu^2 \psi_{\sigma} \sum_{\tau=0}^t (\tilde{\eta}^\tau)^2 \prod_{k=\tau+1}^t ((1+\beta)^2  + (\eta_g^t)^2 \nu^2 \psi+\beta^2) \\
& \leq \nu^2 \psi_{\sigma} \sum_{\tau=0}^t (\tilde{\eta}^\tau)^2 \prod_{k=\tau+1}^t (1 + 2\beta(\beta+1)  + (\eta_g^t)^2 \nu^2\psi) \\
& \leq \nu^2 \psi_{\sigma} \sum_{\tau=0}^t (\tilde{\eta}^\tau)^2 \prod_{k=\tau+1}^t 2\beta(\beta+1) \cdot \prod_{k=\tau+1}^t(1  + (\eta_g^t)^2 \nu^2\psi) \\
& \leq \nu^2 \psi_{\sigma} \sum_{\tau=0}^t (\tilde{\eta}^\tau)^2 \cdot (2\beta(\beta+1))^{t-\tau} \cdot \prod_{k=\tau+1}^t \exp\{(\eta_g^t)^2 \nu^2\psi\}, \\
\end{aligned}
$$
where the last two inequalities uses $\prod_{t=0}^T (a + b) \leq \prod_{t=0}^T a \cdot \prod_{t=0}^T$ if $a>0, b>0, (a-1)(b-1) \geq 1$ and $1+a \leq \exp\{a\}, \forall a>0$.

Analogous to previous analysis, let $\tilde{\eta}^t \leq \eta_g^t \leq \sqrt{\frac{c}{t}}$ for some $c>0$ to obtain
$$
\begin{aligned}
\mathbb{E}\|\bs{x}^{t+1} - \tilde{\bs{x}}^{t+1}\|^2 & \leq \nu^2 \psi_{\sigma} \sum_{\tau=0}^t  (2\beta(\beta+1))^{t-\tau} \cdot \exp\{\nu^2c\psi \sum_{k=\tau+1}^t \frac{1}{k}\} \cdot \frac{(\tilde{\eta}^\tau)^2}{(\tilde{\eta}^t)^2} \cdot (\tilde{\eta}^t)^2 \\
& \leq \nu^2 \psi_{\sigma} \sum_{\tau=0}^t (2\beta(\beta+1))^{t-\tau} \cdot \exp\{\nu^2c\psi \log(\frac{t}{\tau})\} \cdot \frac{t}{\tau} \cdot (\tilde{\eta}^t)^2 \\
& = \nu^2 \psi_{\sigma} \sum_{\tau=0}^t (2\beta(\beta+1))^{t-\tau} \cdot \exp\{\nu^2c\psi \log(\frac{t}{\tau})\} \cdot \frac{t}{\tau} \cdot (\tilde{\eta}^t)^2\\
& = \nu^2 \psi_{\sigma} \sum_{\tau=0}^t (2\beta(\beta+1))^{t-\tau} \cdot  (\frac{t}{\tau})^{1+\nu^2c\psi}\cdot (\tilde{\eta}^t)^2 \\
& \leq \nu^2 \psi_{\sigma} t^{1+\nu^2c\psi} \frac{(2\beta(\beta+1))^{t+1} - 1}{2\beta(\beta+1)-1} \cdot \sum_{\tau=0}^t (\frac{1}{\tau})^{1+\nu^2c\psi}\cdot (\tilde{\eta}^t)^2 \\
& \leq \frac{\psi_{\sigma}}{c\psi} \frac{(2\beta(\beta+1))^{t+1} - 1}{2\beta(\beta+1)-1} t^{1+\nu^2c\psi} (\tilde{\eta}^t)^2. \\
\end{aligned}
$$
In short, we have
\begin{equation}\label{fedavgm:stab_t2}
\begin{aligned}
\frac{1}{T}\sum_{t=0}^{T-1} \mathbb{E}\| \bs{x}^t - \tilde{\bs{x}}^t\|^2 & \leq \mathbb{E}\|\bs{x}^{T} - \tilde{\bs{x}}^{T}\|^2 \leq \Psi_{\text{stab}} \cdot (\tilde{\eta}^t)^2, 
\end{aligned}
\end{equation}
where we let $\Psi_{\text{stab}} = \frac{\psi_{\sigma}}{c\psi} \psi_{\beta} T^{1+\nu^2c\psi}$ and $\psi_{\beta} = \frac{(2\beta(\beta+1))^{T} - 1}{2\beta(\beta+1)-1}$ denotes the scale effects from momentum parameter $\beta$. Moreover, we note that $0< \beta < 1$, which induces that $(1+\beta)^{T}$ dominates the $\psi_{\beta}$.

\subsection{Convergence Analysis} 
We recall the updated rule of the global model as 
$$
\bs{x}^{t+1} = \bs{x}^{t} - \eta_g \bs{m}^{t}, \quad \bs{m}^{t} = \beta \bs{m}^{t-1} + \nu \bs{d}^{t},
$$
where $\bs{d}^{t} = \frac{1}{N} \sum_{i=1}^N \bs{d}_i^{t}$.
For simplicity, we consider the regularized momentum as:
$$
\bs{x}^{t+1} = \bs{x}^{t} - \eta \tilde{\bs{m}}^{t} = \bs{x}^{t} - \eta \frac{1}{N} \sum_{i=1}^N \frac{\tilde{\bs{m}}_i^t}{K},
$$
where 
$$
\tilde{\bs{m}}_i^t = \beta \tilde{\bs{m}}_i^{t-1} + \nu \tilde{\bs{d}}_i^{t}, \quad \tilde{\bs{d}}_i^{t} = \sum_{k=1}^K \bs{g}_i^{t, k-1},
$$ 
denotes the local momentum of the $i$-th client and $\eta = K \eta_l \eta_g$. We note that $\tilde{\bs{m}}^{t} = \frac{1}{\eta_l K} \bs{m}^{t}$. Moreover, we unroll the recursion of $\bs{m}^{t}$ for $t = 0, \dots, T$ with $\bs{m}^{0} = \nu \tilde{\bs{d}}^0$ to obtain
$$
\tilde{\bs{m}}^t = \nu \sum_{\tau=0}^t \beta^{t-\tau} \tilde{\bs{d}}^\tau.
$$

Analogous to \eqref{eq:descent}, we have similar descent lemma with momentum $\bs{m}^t$ as
\begin{equation}\label{eq:fosm_convergence}
\begin{aligned}
\mathbb{E}\left[f(\bs{x}^{t+1})\right] & \leq f(\bs{x}^t) - \frac{\eta}{2} \mathbb{E}\|\nabla f(\bs{x}^t)\|^2 +  \frac{\eta}{2}\underbrace{\mathbb{E}\left[\| \nabla f(\bs{x}^t) - \tilde{\bs{m}}^t\|^2\right]}_{T_1} + \frac{L}{2}\eta^2 \underbrace{\mathbb{E}\left[\|\tilde{\bs{m}}^t\|^2\right]}_{T_2}.
\end{aligned}
\end{equation}

\textbf{Bounding $T_1$.} Using the definition of momentum, we know
$$
\begin{aligned}
& \quad \mathbb{E}\left[\| \nabla f(\bs{x}^t) - \tilde{\bs{m}}^t\|^2\right] \\
& =\mathbb{E}\left\|\nabla f(\bs{x}^t) - \beta\tilde{\bs{m}}^{t-1}  - \nu \tilde{\bs{d}}^t\right\|^2 \quad \quad \quad \triangleright \text{Let $\nu=1-\beta$}\\
& = \mathbb{E}\left\|\beta(\nabla f(\bs{x}^t) - \nabla f(\bs{x}^{t-1})) +\beta(\nabla f(\bs{x}^{t-1}) - \tilde{\bs{m}}^{t-1})  +(1-\beta) (\nabla f(\bs{x}^t) - \tilde{\bs{d}}^t)\right\|^2 \\
& \leq \beta^2 W \mathbb{E}\|\nabla f(\bs{x}^t) - \nabla f(\bs{x}^{t-1})\|^2 + \beta^2 \mathbb{E}\|\nabla f(\bs{x}^{t-1}) - \tilde{\bs{m}}^{t-1}\|^2 + (1-\beta)^2 \mathbb{E}\| \nabla f(\bs{x}^t) - \tilde{\bs{d}}^t\|^2 \\
& \leq \beta^2 W L^2 \mathbb{E}\|\bs{x}^t - \bs{x}^{t-1}\|^2 + \beta^2\mathbb{E}\|\nabla f(\bs{x}^{t-1}) - \tilde{\bs{m}}^{t-1}\|^2 + (1-\beta)^2 \mathbb{E}\|\nabla f(\bs{x}^t) - \tilde{\bs{d}}^t\|^2 \\
& \leq \beta^2 \mathbb{E}\|\nabla f(\bs{x}^{t-1}) - \tilde{\bs{m}}^{t-1}\|^2 + \beta^2 L^2W\eta^2\mathbb{E}\|\tilde{\bs{m}}^{t-1}\|^2 + \nu^2 \mathbb{E}\|\nabla f(\bs{x}^t) - \tilde{\bs{d}}^t\|^2 \\
\end{aligned}
$$

In the above equations, we suppose a constant $W$ that always holds the third equation for all $t$. Then, we can shrink the global learning to absorb the scaling constant $W$, i.e., $\eta^2 \leq W\eta^2$ in the last equation. 

Then, unrolling the recursion from time $t$ down to $0$ with edge conditions $\mathbb{E}\|\nabla f(\bs{x}^{-1}) - \tilde{\bs{m}}^{-1}\|^2 = \mathbb{E}\|\tilde{\bs{m}}^{-1}\|^2 = 0$, we obtain 
\begin{equation}\label{eq:t1_fosm}
\begin{aligned}
\mathbb{E}\| \nabla f(\bs{x}^t) - \tilde{\bs{m}}^t\|^2 & \leq \beta^2 L^2\eta^2 \sum_{\tau=0}^{t-1} \beta^{2(t-\tau)} \mathbb{E}\|\tilde{\bs{m}}^{\tau}\|^2 + \nu^2 \sum_{\tau=0}^{t-1} \beta^{2(t-\tau)} \underbrace{\mathbb{E}\|\nabla f(\bs{x}^{\tau}) - \tilde{\bs{d}}^{\tau}\|^2}_{\text{Bounded by \eqref{eq:local_divergence}}} \\
& \leq \beta^2 L^2\eta^2 \underbrace{\sum_{\tau=0}^{t-1} \beta^{2(t-\tau)} \mathbb{E}\|\tilde{\bs{m}}^{\tau}\|^2}_{\text{Bounded by \eqref{eq:temp_bounded_cumulated_momentum}}} \\
& \quad + \nu^2 \beta^2 \frac{1-\beta^{2t}}{1-\beta^2} \left(\frac{15}{16}\mathbb{E}\|\nabla f(\bs{x}^t)\|^2 + 60L^2\left(9\sigma_l^2 + K\sigma_g^2\right)K\eta_l^2\right).
\end{aligned}
\end{equation}

\textbf{Bounding $T_2$.} 
Given the definition of $\tilde{\bs{m}}^t = \nu \sum_{\tau=0}^t \beta^{t-\tau} \tilde{\bs{d}}^\tau$, we know
\begin{equation}\label{eq:t2_fosm}
\begin{aligned}
\mathbb{E}\|\tilde{\bs{m}}^{t}\|^2 & = \mathbb{E}\|\nu \sum_{\tau=0}^t \beta^{t-\tau} \tilde{\bs{d}}^\tau\|^2 \leq  \nu^2 \mathbb{E}\|\sum_{\tau=0}^t \beta^{t-\tau} \tilde{\bs{d}}^\tau\|^2 \leq \nu^2 (t+1) \sum_{\tau=0}^t \beta^{2(t-\tau)} \mathbb{E}\| \tilde{\bs{d}}^\tau\|^2 \\
& \leq \nu^2 (t+1) \sum_{\tau=0}^t \beta^{2(t-\tau)} \frac{1}{\eta_l^2 K^2}\left( 5 K\eta_l^2 (\sigma_l^2 + 6K\sigma_g^2) + 30K^2\eta_l^2\mathbb{E}\left\| \nabla f(\bs{x}^t) \right\|^2 \right) \quad\quad\quad \triangleright \text{Lemma~\ref{lemma:local_drift}}\\
& \leq \nu^2 (t+1) \frac{1-\beta^{2(t+1)}}{1-\beta^2} \frac{1}{\eta_l^2 K^2} \left( 5 K\eta_l^2 (\sigma_l^2 + 6K\sigma_g^2) + 30K^2\eta_l^2\mathbb{E}\left\| \nabla f(\bs{x}^t) \right\|^2 \right). \\
\end{aligned}
\end{equation}

Based on the above results, we bound the cumulated momentum in $T_1$:
\begin{equation}\label{eq:temp_bounded_cumulated_momentum}
\begin{aligned}
& \quad \sum_{\tau=0}^{t-1} \beta^{2(t-\tau)} \mathbb{E}\|\tilde{\bs{m}}^{\tau}\|^2 \\
& \leq \frac{\nu^2}{1-\beta^2} \sum_{\tau=0}^{t-1} \beta^{2(t-\tau)} (\tau+1)(1-\beta^{2(\tau+1)}) \frac{1}{\eta_l^2 K^2} \left( 5 K\eta_l^2 (\sigma_l^2 + 6K\sigma_g^2) + 30K^2\eta_l^2\mathbb{E}\left\| \nabla f(\bs{x}^t) \right\|^2 \right) \\
& \leq \frac{\nu^2}{1-\beta^2} \sum_{\tau=0}^{t-1} \beta^{2(t-\tau)} (\tau+1) \frac{1}{\eta_l^2 K^2} \left( 5 K\eta_l^2 (\sigma_l^2 + 6K\sigma_g^2) + 30K^2\eta_l^2\mathbb{E}\left\| \nabla f(\bs{x}^t) \right\|^2 \right) \\
& \leq \frac{\nu^2}{1-\beta^2} (t+1) \frac{1-\beta^{2(t+1)}}{1-\beta^2} \frac{1}{\eta_l^2 K^2} \left( 5 K\eta_l^2 (\sigma_l^2 + 6K\sigma_g^2) + 30K^2\eta_l^2\mathbb{E}\left\| \nabla f(\bs{x}^t) \right\|^2 \right).
\end{aligned}
\end{equation}

\textbf{Putting together.}
Substituting \eqref{eq:temp_bounded_cumulated_momentum} to \eqref{eq:t1_fosm} to obtain the bound of $T_1$, and then combining $T_1$, \eqref{eq:t2_fosm} and \eqref{eq:fosm_convergence}, we derive the the decent lemma:
\begin{align}
\mathbb{E}\|\nabla f(\boldsymbol{x}^t)\|^2 & \leq \frac{2(f(\boldsymbol{x}^t) - \mathbb{E}\left[f(\boldsymbol{x}^{t+1})\right])}{\eta}  \nonumber \\
& \quad + L^2 \eta^2 \frac{\nu^2\beta^2}{1-\beta^2} (t+1) \frac{1-\beta^{2(t+1)}}{1-\beta^2} \frac{1}{\eta_l^2 K^2} \left( 5 K\eta_l^2 (\sigma_l^2 + 6K\sigma_g^2) + 30K^2\eta_l^2\mathbb{E}\left\| \nabla f(\boldsymbol{x}^t) \right\|^2 \right) \label{eq:scale1}\\
& \quad + \nu^2\beta^2 \frac{1-\beta^{2t}}{1-\beta^2} \left(\frac{15}{16}\mathbb{E}\|\nabla f(\boldsymbol{x}^t)\|^2 + 60L^2\left(9\sigma_l^2 + K\sigma_g^2\right)K\eta_l^2\right) \nonumber \\ 
& \quad + L \eta \nu^2 (t+1) \frac{1-\beta^{2(t+1)}}{1-\beta^2} \frac{1}{\eta_l^2 K^2} \left( 5 K\eta_l^2 (\sigma_l^2 + 6K\sigma_g^2) + 30K^2\eta_l^2\mathbb{E}\left\| \nabla f(\boldsymbol{x}^t) \right\|^2 \right). \label{eq:scale2}
\end{align}

Aligning with the previous analysis, we denote $\tilde{\eta} = \eta_l \eta_g$, $\eta_g \leq \sqrt{\frac{c}{T}}$, $\eta_l \leq \frac{1}{\sqrt{T} K L}$, which yields $\eta_l\eta_g \leq \frac{\tilde{c}}{(t+1)L}$ for a constant $\tilde{c}>0$ and $t\in[T-1]$ to absorb the coefficient $t$ in \eqref{eq:t2_fosm}. Specifically, we scale $\tilde{\eta}L(t+1)$ to $\tilde{c}$ for \eqref{eq:scale1} and $\tilde{\eta}L(t+1)$ to $\tilde{\eta}L \tilde{c}$ for \eqref{eq:scale2}. Then, we reorganize the terms to have
$$
\begin{aligned}
\rho \mathbb{E}\|\nabla f(\boldsymbol{x}^t)\|^2 & \leq \frac{2(f(\boldsymbol{x}^t) - \mathbb{E}\left[f(\boldsymbol{x}^{t+1})\right])}{K \tilde{\eta}} + \Psi_1 \tilde{\eta} + \Psi_2
,\end{aligned}
$$
where
$$
\begin{aligned}
\rho & = 1- \frac{15}{16}\left(1 + 32 \tilde{c}(1+\frac{1}{\beta^2})L\tilde{\eta} \right) \cdot \frac{\nu^2\beta^2}{1-\beta^2} (1-\beta^{2(t+1)}) > 0\\
\Psi_1 & = (1-\beta^{2(t+1)}) \frac{\tilde{c}\nu^2}{(1-\beta^2)^2} \cdot 5L (\sigma_l^2 + 6K\sigma_g^2), \\
\Psi_2 & =  (1-\beta^{2t}) \frac{\nu^2\beta^2}{1-\beta^2} \cdot 60L^2 \left(9\sigma_l^2 + K\sigma_g^2\right)K\eta_l^2
\end{aligned}
$$

Then, taking the average of the above equation from time $0$ to $T-1$ and full expectation, we have
\begin{equation}\label{fedavgm:conv_t1}
\begin{aligned}
\frac{1}{T}\sum_{t=0}^{T-1} \mathbb{E}\|\nabla f(\bs{x}^t)\|^2 \leq \frac{2D}{\hat{\rho} T K \tilde{\eta}} + \frac{1}{T}\sum_{t=0}^{T-1} \frac{\Psi_1}{\hat{\rho}} \tilde{\eta} + \frac{1}{T}\sum_{t=0}^{T-1} \frac{\Psi_2}{\hat{\rho}},
\end{aligned}
\end{equation}
where 
$\hat{\rho} = \min_{t} \rho$. 

\subsection{Excess Risk Analysis}

Now, we provide the excess risk upper bound of FOSM methods. Analogous to our previous analysis, we have
\begin{equation}\label{eq:excess_target_fedavgm}
\begin{aligned}
\min_{t\in[T]} \mathcal{E}(\bs{x}^t) \leq \frac{1}{T}\sum_{t=0}^{T-1} \mathcal{E}(\bs{x}^t) & \leq \phi_1 \cdot \underbrace{\frac{1}{T}\sum_{t=0}^{T-1} \mathbb{E}\| \bs{x}^t - \tilde{\bs{x}}^t\|^2}_{\text{stability dynamics}} +  \phi_2 \cdot \underbrace{\frac{1} {T}\sum_{t=0}^{T-1} \mathbb{E}\|\nabla f(\bs{x}^t)\|^2}_{\text{gradient dynamics}}. \\
\end{aligned}
\end{equation}

Then, substituting corresponding terms with \eqref{fedavgm:conv_t1} and \eqref{fedavgm:stab_t2}, we have
$$
\begin{aligned}
\min_{t\in[T]} \mathcal{E}(\bs{x}^t) \leq \underbrace{\frac{\tilde{D}}{\tilde{\eta} T} + \bar{\Psi}_1 \tilde{\eta} +  \bar{\Psi}_{\text{stab}} \tilde{\eta}^2}_{\text{stability and convergence  trade-off}} + \underbrace{\frac{1}{T}\sum_{t=0}^{T-1} \frac{\Psi_2}{\hat{\rho}}}_{\text{client drift error}},
\end{aligned}
$$
where 
$$
\tilde{D} = \frac{2\phi_1 D}{\hat{\rho}K}, \quad \bar{\Psi}_1 = \frac{1}{T}\sum_{t=0}^{T-1} \frac{\phi_1 \Psi_1}{\hat{\rho}},\quad \bar{\Psi}_{\text{stab}} = \phi_2 \frac{\psi_{\sigma}}{c\psi} \psi_{\beta} T^{1+\nu^2c\psi}.
$$ 

Using $\eta_g \leq \sqrt{\frac{c}{T}}$ and Lemma~\ref{lemma:constant_stepsize} with $d = \sqrt{\frac{T}{c}}, r_0 = \tilde{D}, b = \bar{\Psi}_1$, and $e=\Psi_{\text{stab}}$, we obtain
\begin{equation}
\min_{t\in[T]} \mathcal{E}(\bs{x}^t) \leq 2 \left(\frac{\bar{\Psi}_1 \tilde{D}}{T}\right)^{\frac{1}{2}}+2 \bar{\Psi}_{\text{stab}}^{\frac{1}{3}}\left(\frac{\tilde{D}}{T}\right)^{\frac{2}{3}}+\frac{\tilde{D}}{\sqrt{Tc}} + \underbrace{\frac{1}{T}\sum_{t=0}^{T-1} \frac{\Psi_2}{\hat{\rho}}}_{\text{client drift error}} .
\end{equation}

Noting the rate of each term satisfies:
$$
\begin{aligned}
\bar{\Psi}_1 = \mathcal{O}\left((1-\beta^{T}) \cdot \sigma_K^2\right), \Psi_{\text{stab}} = \mathcal{O}\left((1+\beta)^{T} K\sigma_n^2 T^{1+\nu^2 c \psi})\right),
\end{aligned}
$$
we obtain the bound of the minimum excess risk:
\begin{equation}
\begin{aligned}
\min_{t\in[T]} \mathcal{E}(\bs{x}^t) & \leq \mathcal{O}\left(\sqrt{(1-\beta^T)\cdot\frac{\sigma_K^2 D}{KT}}\right) + \mathcal{O}\left((1-\beta^T)\cdot \frac{\sigma_K^2}{T}\right) \\
& \quad + \mathcal{O}\left(\left((1+\beta)^T\cdot\frac{\sigma_n^2 D^2}{Kc}\right)^{\frac{1}{3}} \cdot \left(\frac{1}{T}\right)^{\frac{1-\nu^2c\psi}{3}} + \frac{D}{K\sqrt{Tc}}\right),
\end{aligned}
\end{equation}
which concludes the proof.

\section{E. Additional Experiment Results}\label{app:exp}

Our implementation is based on FL framework FedLab~\citep{JMLR:v24:22-0440}. Our experiments are conducted on a computation platform with NVIDIA 2080TI GPU * 2.

\begin{figure}[t]
\centering
\subfigure[Batch size $b=16$]{
\begin{minipage}[t]{\textwidth}
\centering
\includegraphics[width=\linewidth]{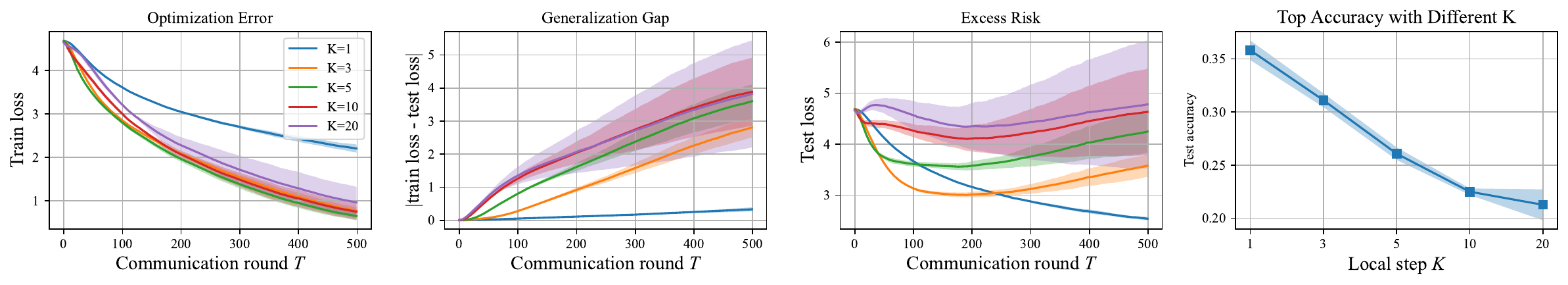}
\end{minipage}}
\subfigure[Batch size $b=32$]{
\begin{minipage}[t]{\textwidth}
\centering
\includegraphics[width=\linewidth]{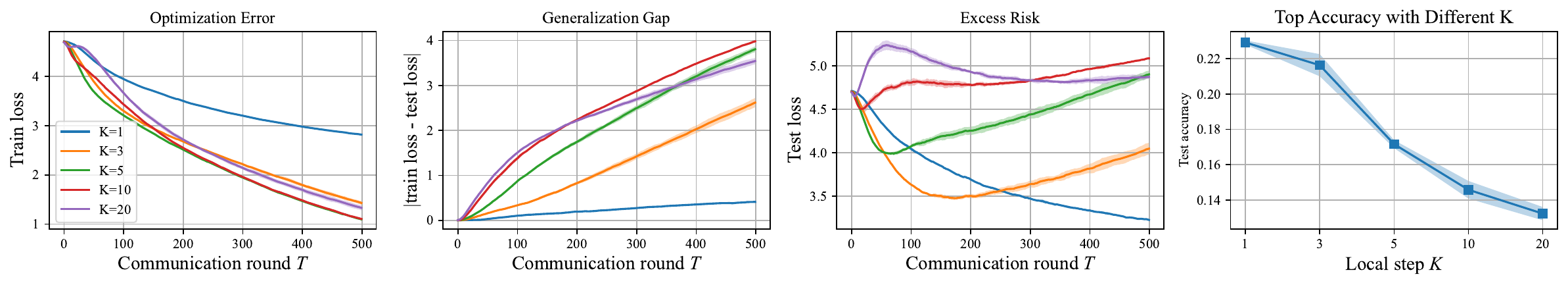}
\end{minipage}}
\hspace{0.5mm}
\subfigure[Batch size $b=64$]{
\begin{minipage}[t]{\textwidth}
\includegraphics[width=\linewidth]{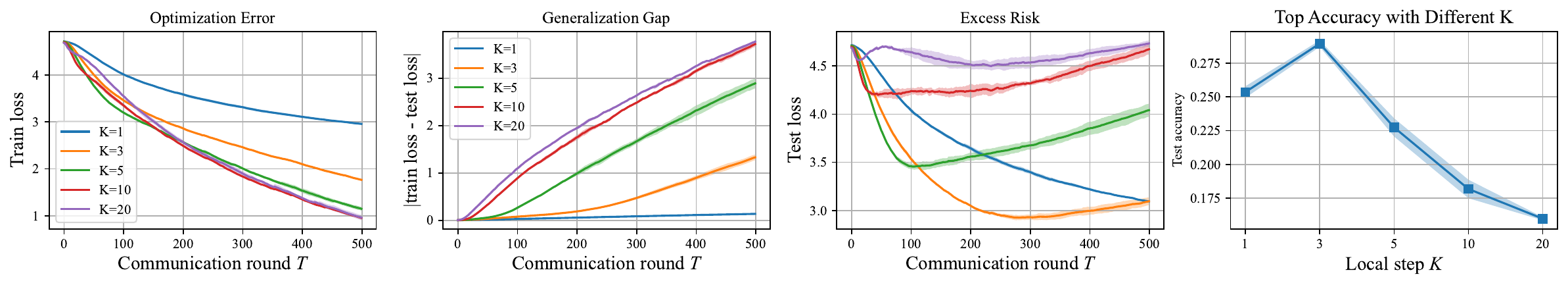}
\end{minipage}}
\hspace{0.5mm}
\caption{Proof-of-concept on CIFAR100 experiments with different local batch size $b$.}\label{fig:cifar100}
\end{figure}

\textbf{Experiments on CIFAR-100 and ResNet18-GN.}\quad 
We present the results of CIFAR-100 experiments in Figure~\ref{fig:cifar100} with the same setting described in the main paper. The training loss curve decreases faster with a larger $K$. According to the generalization gap, the generalization of the CIFAR-100 task is more sensitive to the selection of $K$. Especially when the batch size is relatively smaller, larger $K$ leads to easily overfitting. 

\begin{figure}[t]
\centering
\subfigure[Batch size $b=16$]{
\begin{minipage}[t]{\textwidth}
\centering
\includegraphics[width=\linewidth]{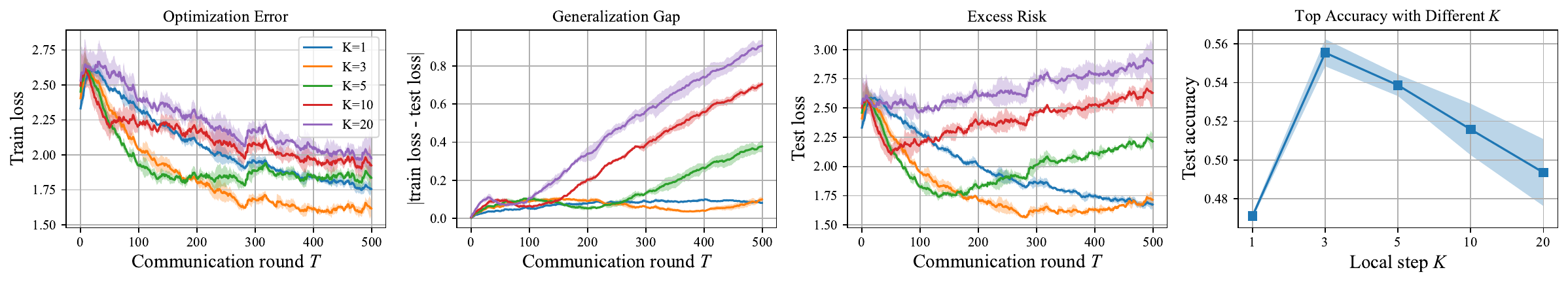}
\end{minipage}}
\hspace{0.5mm}
\subfigure[Batch size $b=64$]{
\begin{minipage}[t]{\textwidth}
\includegraphics[width=\linewidth]{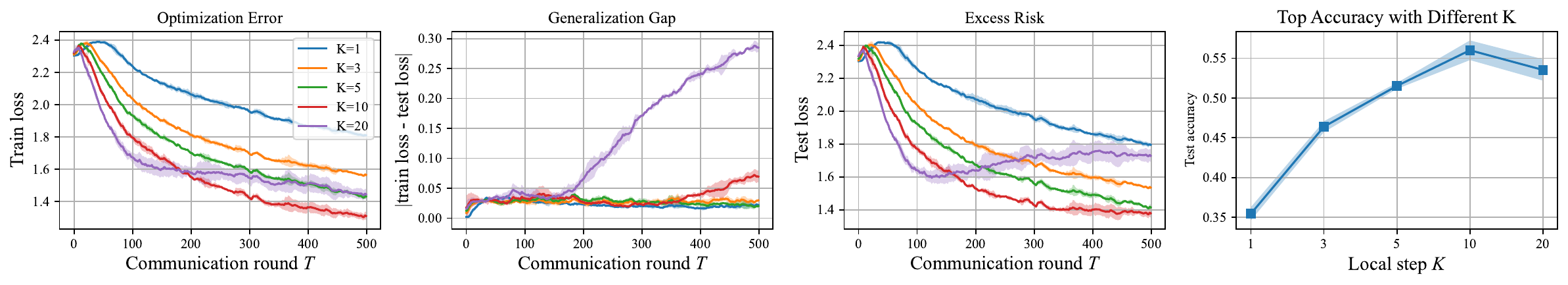}
\end{minipage}}
\hspace{0.5mm}
\caption{Proof-of-concept on CIFAR10 experiments with different local batch size $b$.}\label{app:exp:cifar10}
\end{figure}

\textbf{Sensitivity study on batch size.}\quad
For both CIFAR-10 and CIFAR100 task, we select batch size $b = \{16, 32, 64\}$ and run the experiments. In the main paper, we reported the results of CIFAR-10 with $b=32$. Here, we present missing results in Figure~\ref{app:exp:cifar10} for the CIFAR-10 task. The optimal $K$ of the CIFAR-10 task follows our previous analysis, that is, optimal $K$ is proportional to batch size $K$. In CIFAR100 task, we find that the optimal $K=1$ happens low batch size $b=\{16, 32\}$. The optimal $K=3$ happens in a relatively large batch size of $b=64$. It matches our findings on the CIFAR-10 task, where optimal $K$ is proportional to batch size $b$. This is because batch size affects local stochastic gradient variance $\sigma_l^2$. 

\begin{figure}[t]
\centering
\subfigure[Batch size $b=16$, local steps $K=3$, and learning rate decay $\epsilon$\label{app:exp:cifar100_lrd}]{
\begin{minipage}[t]{\textwidth}
\centering
\includegraphics[width=\linewidth]{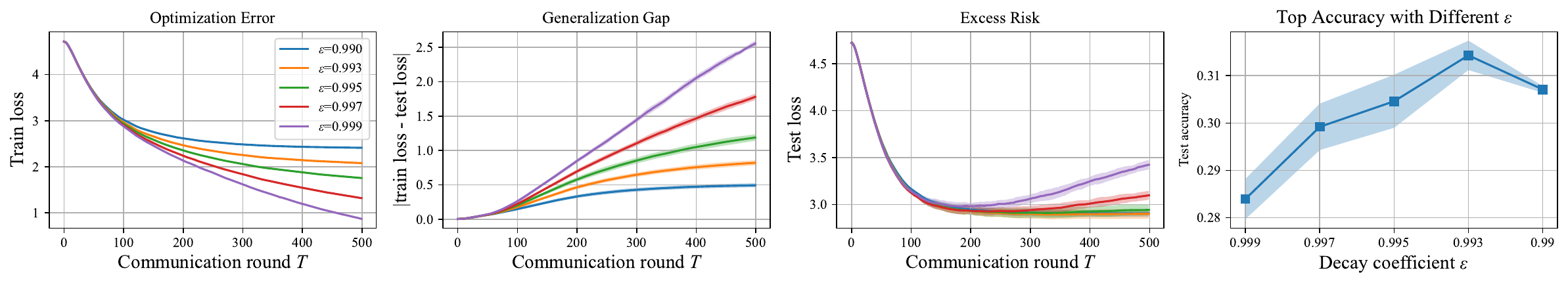}
\end{minipage}}
\hspace{0.5mm}
\subfigure[Batch size $b=16$, local steps $K=1$, and server momentum $\beta$\label{app:exp:cifar100_mom}]{
\begin{minipage}[t]{\textwidth}
\includegraphics[width=\linewidth]{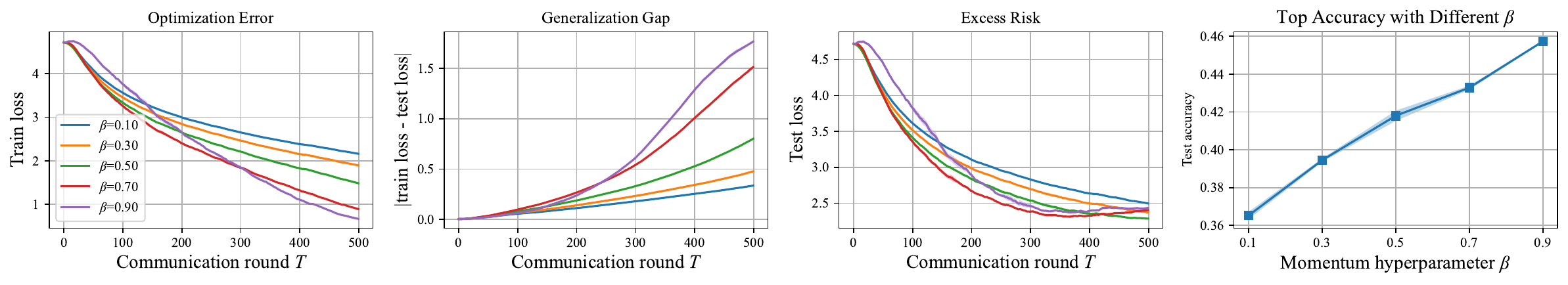}
\end{minipage}}
\hspace{0.5mm}
\caption{Proof-of-concept on CIFAR100 experiments with global learning rate decay and momentum.}\label{app:exp:cifar100_mom_lrd}
\end{figure}

\textbf{Global learning rate decay evaluation on CIFAR100 setting.}\quad
We apply global learning rate decay on the CIFAR-100 setting as described in the main paper. We select a setting with $b=16$ and $K=3$ in Figure~\ref{app:exp:cifar10} where the model starts over-fitting after 150 rounds. As shown in Figure~\ref{app:exp:cifar100_lrd}, global learning rate decay helps the model generalize stably and better. In particular, we observe that test curves with decay coefficient $\beta = \{0.99, 0.995, 0.997\}$ convergence stably, and the overfitting phenomenon is resolved. The results match our theories and discussion in the main paper.

\textbf{Server momentum evaluation on CIFAR100 setting.}\quad 
We run FOSM on the CIFAR-100 setting as shown in Figure~\ref{app:exp:cifar100_mom}. Since we intend to observe the generalization performance of FOSM in comparison with Algorithm~\ref{alg:fedopt}, we choose the setting with hyperparameters local batch size $b=16$ and local step size $K=1$, which obtains the highest test accuracy in Figure~\ref{fig:cifar100}. Importantly, in most cases, FOSM achieves better test accuracy than Algorithm~\ref{alg:fedopt}. Meanwhile, we also observe that a large momentum coefficient typically results in a large generalization gap, which matches our theories in the main paper.

\end{document}